\documentclass[10pt,twocolumn,letterpaper]{article}

\usepackage{iccv}
\usepackage{times}

\usepackage[utf8]{inputenc} % allow utf-8 input
\usepackage[T1]{fontenc}    % use 8-bit T1 fonts
\usepackage{url}            % simple URL typesetting
\usepackage{booktabs}       % professional-quality tables
\usepackage{amsfonts}       % blackboard math symbols
\usepackage{nicefrac}       % compact symbols for 1/2, etc.
\usepackage{microtype}      % microtypography
\usepackage{xcolor}         % colors
\usepackage{multirow}
\usepackage{amsmath}
\usepackage{subcaption}
\usepackage[pdftex]{graphicx} 
\usepackage{caption}
\usepackage{wrapfig}
\usepackage{enumitem}
\usepackage{amsthm}
\usepackage{amssymb}
\usepackage{colortbl}
\newcommand{\ying}[1]{\textcolor{black}{#1}}
\newcommand{\yunqiao}[1]{\textcolor{black}{#1}}
\newcommand{\final}[1]{\textcolor{black}{#1}}
\usepackage[accsupp]{axessibility}
\usepackage{cuted}
\theoremstyle{plain}
\newtheorem{theorem}{Theorem}[section]
\newtheorem{proposition}[theorem]{Proposition}

\theoremstyle{definition}

\theoremstyle{definition}

\usepackage{algorithm}
\usepackage{algorithmic}
\theoremstyle{remark}

\usepackage[textsize=tiny]{todonotes}
\definecolor{aliceblue}{rgb}{0.94, 0.97, 1.0}
\usepackage{appendix}
\usepackage[pagebackref=true,breaklinks=true,letterpaper=true,colorlinks,bookmarks=false]{hyperref}

\iccvfinalcopy % *** Uncomment this line for the final submission

\def\iccvPaperID{3177} % *** Enter the ICCV Paper ID here

\ificcvfinal\fi

\begin{document}

%%%%%%%%% TITLE
\title{Concept-wise Fine-tuning Matters in Preventing Negative Transfer}

\author{
    Yunqiao Yang$^1$\thanks{Part of the work was done when the author interned at Tencent AI Lab.}, Long-Kai Huang$^2$, Ying Wei$^1$\thanks{Corresponding Author} \\
    $^1$City University of Hong Kong  $^2$Tencent AI Lab \\
    {\tt\small \{hustyyq, hlongkai, judyweiying\}@gmail.com}
}

\maketitle
% % Remove page # from the first page of camera-ready.
\ificcvfinal\thispagestyle{empty}\fi

\begin{abstract}
\ying{
A multitude of prevalent pre-trained models mark a major milestone in the development of artificial intelligence, while 
fine-tuning has been a common practice that enables pre-trained models \yunqiao{to} figure prominently in a wide array of target datasets.
Our empirical results reveal that off-the-shelf fine-tuning techniques are far from adequate to mitigate negative transfer caused by two types of underperforming features in a pre-trained model, including rare features and spuriously correlated features.
Rooted in structural causal models of predictions after fine-tuning, we propose a Concept-wise fine-tuning ({Concept-Tuning}) approach which refines feature representations in the level of patches with each patch encoding a concept.
{Concept-Tuning} minimizes the negative impacts of rare features and spuriously correlated features by (1) maximizing the mutual information between examples in the same category with regard to a slice of rare features (a patch) and (2) applying front-door 
adjustment via attention neural networks in channels and feature slices (patches).
The proposed {Concept-Tuning} %remains compatible with prior state-of-the-art fine-tuning methods, and 
consistently and significantly (by up to $4.76\%$) improves prior state-of-the-art fine-tuning methods on \final{eleven} datasets, diverse pre-training strategies (supervised and self-supervised ones), various network architectures, and sample sizes in a target dataset.
}
\end{abstract}

% \vspace{-0.05in}
\section{Introduction}\label{Introduction}
% \vspace{-0.06in}
%
\ying{Pre-trained models, pre-trained by either conventional supervised~\cite{kornblith2019better} or resurgent self-supervised strategies~\cite{grill2020bootstrap,he2020momentum}, undoubtedly constitute a milestone in the artificial intelligence community. Such a resounding success stems from the gap between the heavy reliance of deep neural networks on extensive data on one side and the lack of annotated data in many real-world applications on the other side.}
% A well-established \ying{transfer learning} paradigm, \ying{\ie,} 
Pre-trained models suffice to bridge this gap under the aegis of the well-established fine-tuning paradigm~\cite{he2019rethinking}.
% Fine-tuning \ying{a pre-trained model to downstream tasks~\cite{he2019rethinking}, 
%has been demonstrated to offer strong performance improvements over training downstream tasks from scratch and take the lead to bridge this gap.}
\begin{figure}[t]
  \centering
  \includegraphics[width=\columnwidth]{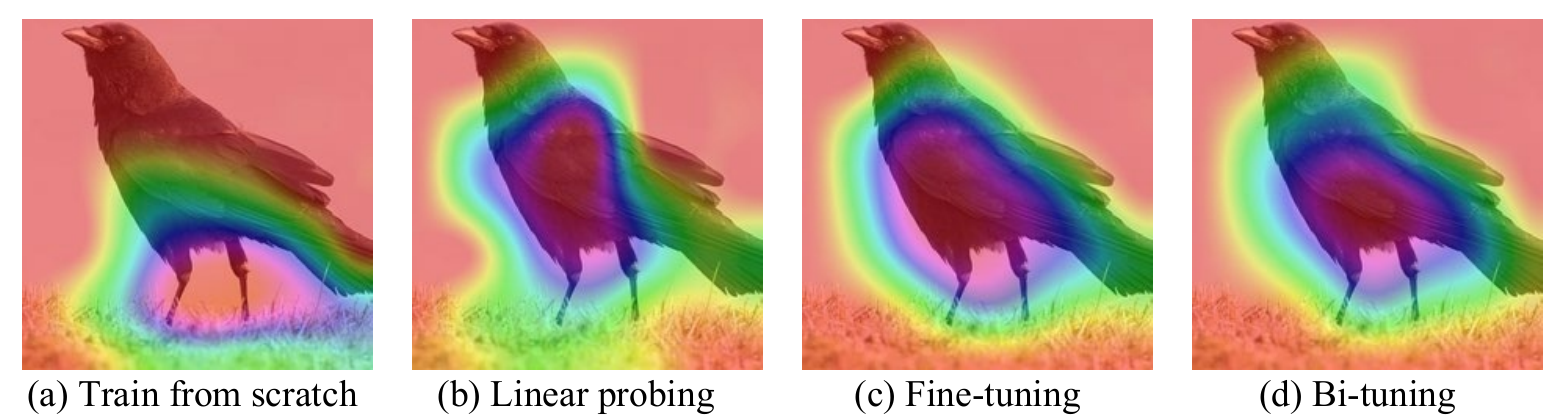}
  \caption{Exemplar attentive regions of the model trained \final{ (a) from scratch, by (b) linear probing, (c) vanilla fine-tuning, and (d) bi-tuning via Eigen-Grad-CAM~\cite{muhammad2020eigen}, where only (a) predicts correctly.}} %, respectively.}
  \label{fig_1}
  \vspace{-0.2in}
\end{figure}

\ying{The practice of fine-tuning, unfortunately but unsurprisingly, is not always outperforming; the notorious problem of \emph{negative transfer}~\cite{chen2019catastrophic, wang2019characterizing} arises} especially when \ying{downstream} tasks are out of the distribution of pre-training data.
%\ying{have very} limited \ying{labeled} data. 
There could be cases \ying{where} the model trained from scratch outperforms fine-tuning. \ying{For example, there exist $3.08\%$ testing images of the downstream dataset CUB~\cite{wah2011caltech} \ying{on} which the model trained from scratch makes correct predictions while fine-tuning the supervised pre-trained model} misclassifies. 
The issue remains even if we resort to one recent state-of-the-art fine-tuning method named Bi-tuning~\cite{zhong2020bi}, despite offering a smaller percentage of $2.74\%$.
%  Ours 2.27%
%the downstream dataset of CUB~\cite{wah2011caltech}, 
%the performances  of  \ying{vanilla} fine-tuning, \ying{the} recent state-of-the-art \ying{of} Bi-tuning~\cite{zhong2020bi},  and  training from scratch  \ying{are $78.01\%$, $82.93\%$, and $54.50\%$, respectively. 
%Despite the superiority of vanilla fine-tuning and Bi-tuning,} 
% there exist $3.08\%$ and $2.74\%$ testing images \ying{on} which the model trained from scratch \ying{makes} correct  \ying{predictions but} \ying{two fine-tuning methods} misclassify.  
\ying{Beyond such a prediction gap between the from-scratch model and fine-tuning models, we do see a gap in attended regions. Fig.~\ref{fig_1} shows that the from-scratch model that predicts correctly pays attention to the feet, but fine-tuning models that misclassify concentrate on body-related features, \final{possibly mislead by ImageNet pre-trained models.}}
% When we look closer to attended regions in one testing image on which the model trained from scratch predicts correctly but both fine-tuning models fail, } 
% % compare the  \ying{regions attended by the three models through the tool of} 
% Fig.~\ref{fig_1} shows that two fine-tuning models both attend body-related features while the from-scratch model pays attention to the feet, which explains the prediction inconsistency.
% \ying{Consistent with the difference in predictions, the two fine-tuning methods rely more on body-related features while the model trained from scratch attends to the feet of the bird.}
\emph{The devil that accounts for such gaps lies in those underperforming features offered by the pre-trained model},
% The devil \ying{that makes such differences} we believe is the  biases incurred by the pre-trained 
% model~\cite{salman2022does}, \ying{, 
among which we focus on the following two types based on our empirical studies.
%for 
%the following two aspects.

\begin{figure}[t]
\vspace{-0.1in}
	\centering
	\begin{subfigure}[b]{\columnwidth}
		\centering
%		\raisebox{0.25\height}{
        {\includegraphics[width =\columnwidth]{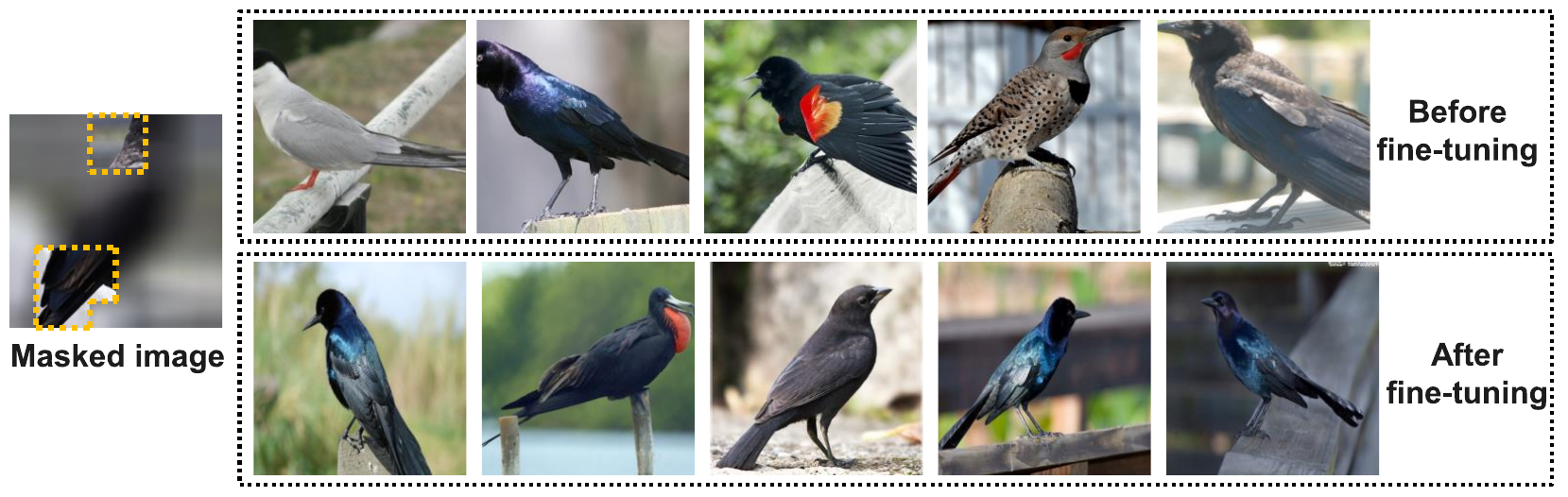}}
        \caption{Top-$5$ similar images \ying{retrieved given a masked image.}}\label{fig:similar_images}
        \end{subfigure}
	% \hfill
	\begin{subfigure}[b]{\columnwidth}
        \centering
        {\includegraphics[width=0.49\columnwidth]{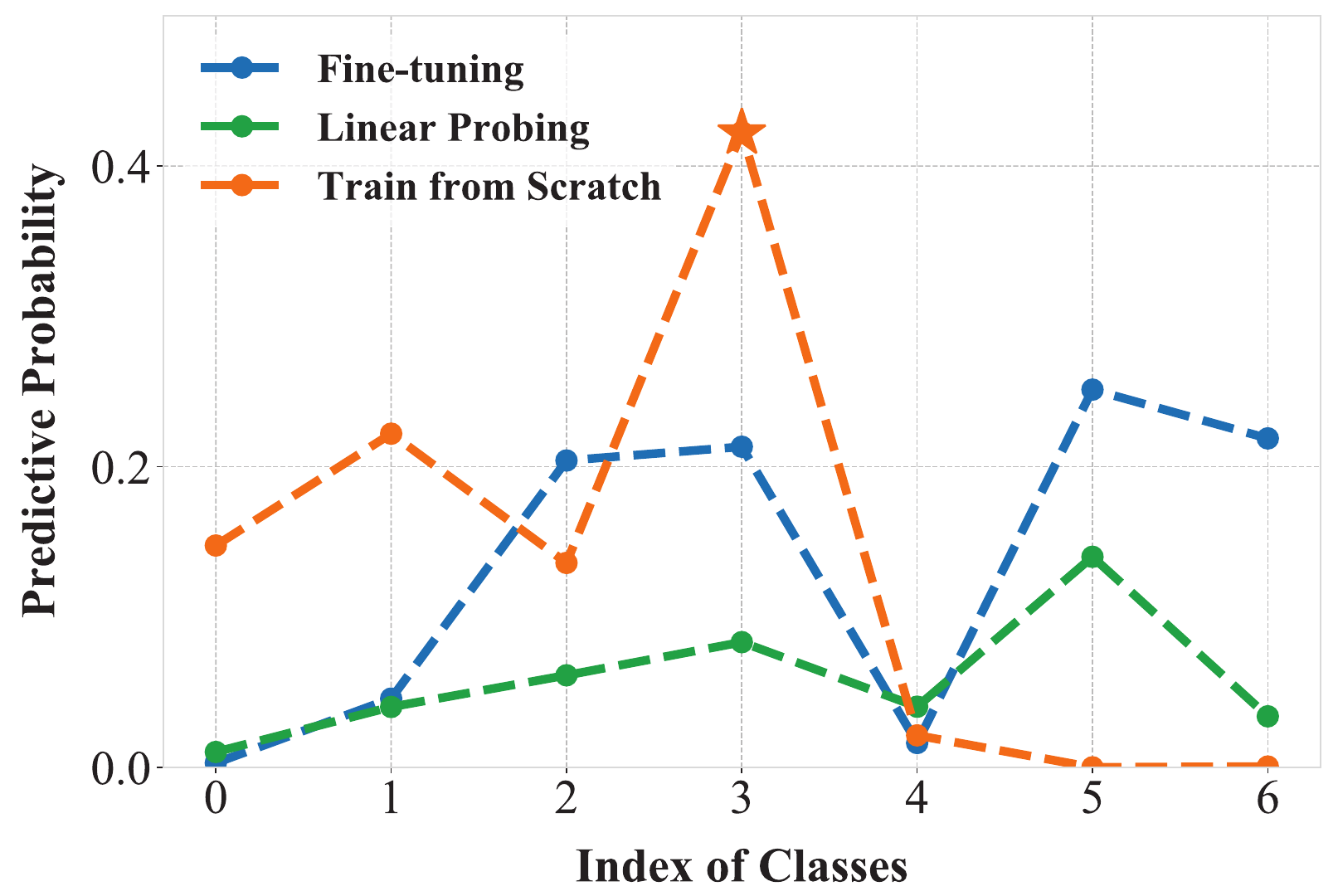}}%}
	\caption{The union of {top-$k$ predictive} probability scores \ying{by different methods, where the star marks the ground-truth class index $3$.}} %across different methods}
	\label{fig:pred_probabilities}
\end{subfigure}

\caption{(a) \ying{Retrieval of} the most similar images \ying{based on the} cosine similarity \ying{between features of the masked image and another one.} (b) Given \ying{the} masked image in (a), the model trained from scratch 
predict\ying{s} correctly \ying{but both fine-tuning and linear probing fail.}}
\label{fig_2}
\vspace{-0.1in}
\end{figure}

\textbf{Type I \emph{(Rare features)}:} These undertrained features by the pre-training dataset force the attention of the fine-tuned model to be diverted to those well-trained features, although the most discriminative features for the downstream dataset are exactly those undertrained ones.

% The features that should be the most discriminative for the downstream dataset, however, is undertrained in the pre-training dataset. These \emph{rare features} force the attention of the model to be diverted to those well-trained but less discriminative features that prevent accurate prediction.
Fig.~\ref{fig:similar_images} demonstrates a slice of rare features, \ie, \final{neck/tail} features. Concretely, 
% To \ying{validate} this issue, 
we %first
mask an original image via Gaussian blurring \ying{with} two patches \ying{preserved}, and 
\ying{retrieve} its most similar images. 
The five most similar images retrieved \final{by neck/tail are unfortunately distant (evidenced by colorful necks instead of the expected dark neck)} using either pre-trained features (before fine-tuning) or fine-tuned features (after fine-tuning), advocating that the fine-grained neck features are undertrained in the coarse-grained pre-training dataset of ImageNet.
As a consequence, both linear probing that only fine-tunes the classification head and vanilla fine-tuning lose the attention of the discriminative neck features, resulting in incorrect predictions, while 
% \ying{As shown in Fig.~\ref{fig:similar_images}, the pre-trained neck features are highly insensitive to color since coarse-grained classification in the pre-training dataset of ImageNet does not depend on fine-grained features like neck color to discriminate; as a consequence,  the neck features after fine-tuning %cannot 
% even struggle to
% differentiate between obviously distant images.
% compared to 
training from scratch where all features are initialized to be unanimously uninformative succeeds.
%, both linear probing that only fine-tunes the classification head and vanilla fine-tuning are susceptible to the rare features. This has been supported by the correct predictive probability of training from scratch but incorrect ones by the other two in Fig.~\ref{fig:pred_probabilities}.

\textbf{Type II~\emph{(Spuriously correlated features)}:}
The misleading correlations in the pre-training dataset constitute another source that diverts the attention of the fine-tuned model.

\begin{figure}[t]
% \vspace{-0.2in}
	\centering
	\begin{subfigure}[]{0.49\columnwidth}
		\centering{
		\includegraphics[width =0.9\columnwidth]{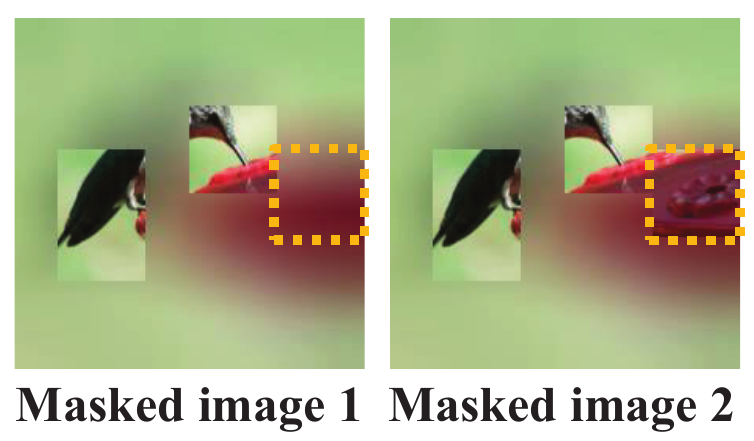}}
%	\caption{Masked images}
	\label{fig:masked_images_spurious}
	\end{subfigure}
	\hfill
	\begin{subfigure}[]{0.49\columnwidth}
		\centering
		 {\includegraphics[width = \columnwidth]{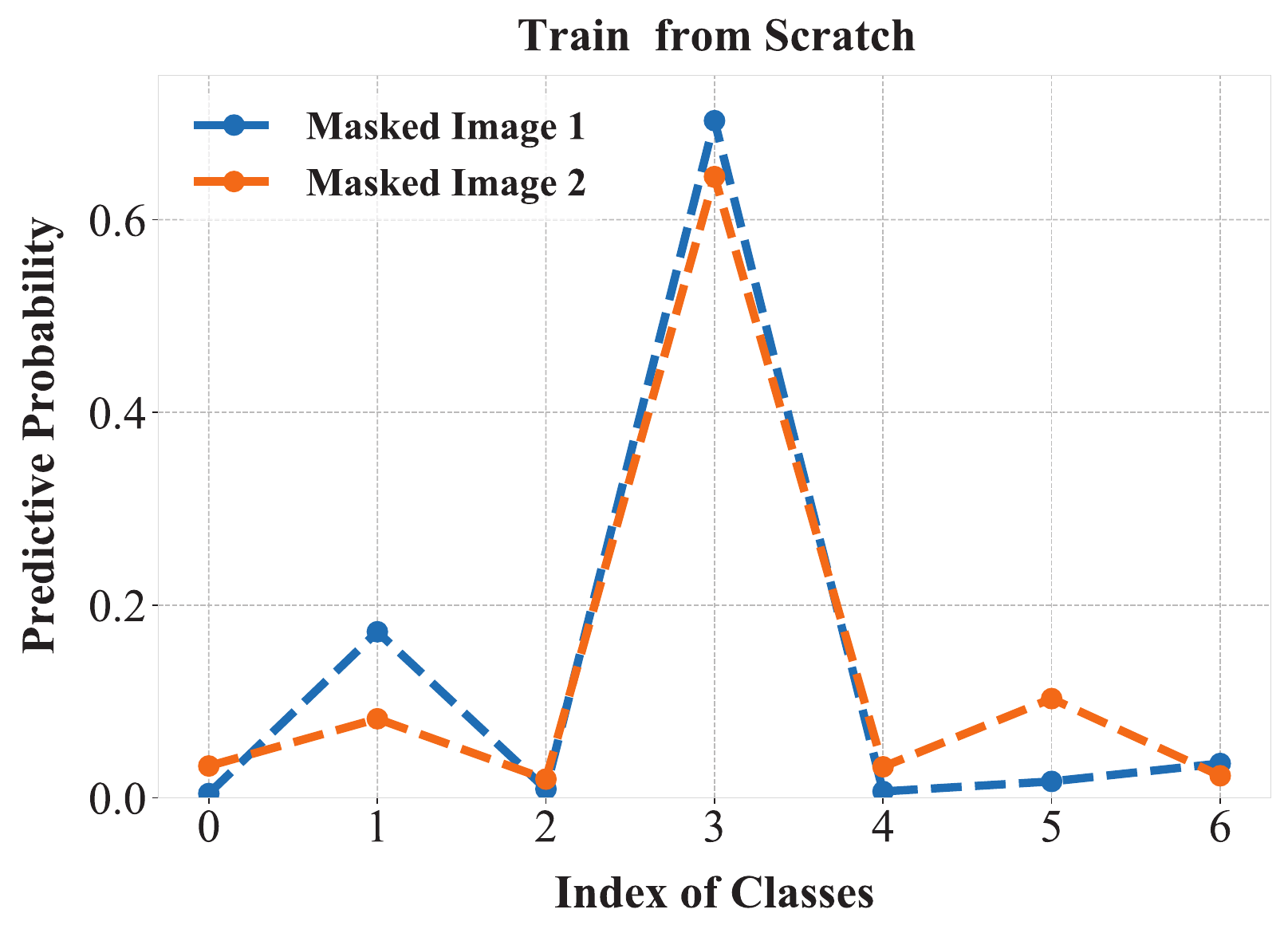}}
%	\caption{Predictive probability of the model trained  from scratch}
	\label{fig:pred_from_scratch}
	\end{subfigure}
        
	\begin{subfigure}[]{0.49\columnwidth}
		\centering
	{\includegraphics[width = \columnwidth]{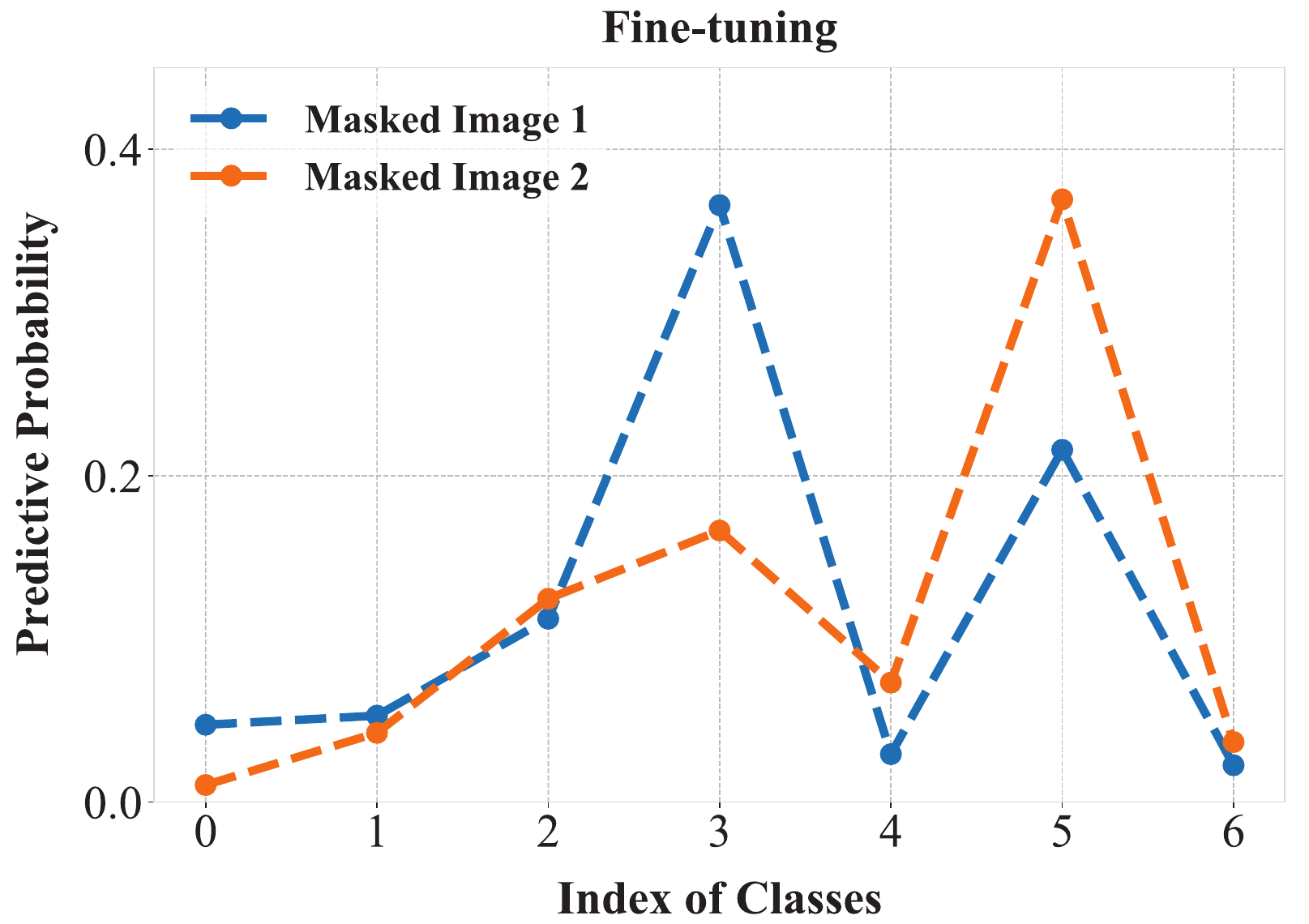}}
%	\caption{Predictive probability of fine-tuning}
	\label{fig:pred_fine-tuning}
	\end{subfigure}
		\hfill
	\begin{subfigure}[]{0.49\columnwidth}
		\centering
		{\includegraphics[width = \columnwidth]{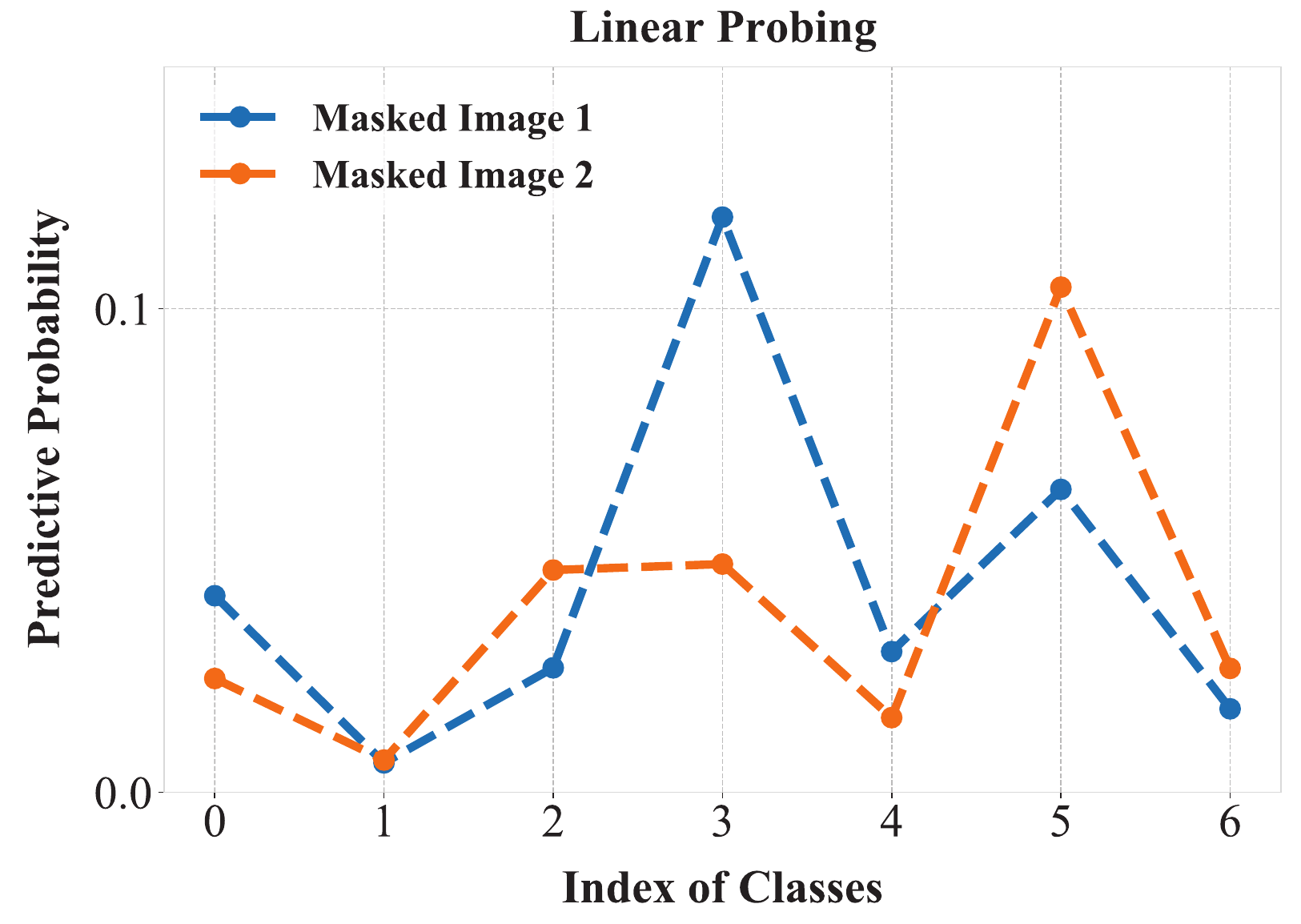}}
%	\caption{Predictive probability of the linear probing}
	\label{fig:pred_linear_probing}
	\end{subfigure}
 \vspace{-0.2in}
\caption{An exampl\ying{ar} pair of spuriously correlated features\ying{, where the model trained from scratch predicts consistently after including one more patch in the masked image 2. Unfortunately, fine-tuning and linear probing predict incorrectly with the introduction of the same patch.}}
%. Given the two masked images with only one patch difference (as shown in the red box), the model trained from scratch can predict well on both masked images, while the performances of fine-tuning and linear probing become worse when providing one more patch, which we define as spuriously correlated features}
\label{fig_3}
\vspace{-0.1in}
\end{figure}

In Fig.~\ref{fig_3}, we show head features and bird feeder features as a pair of spuriously correlated features. For the masked image 1 with only the two patches of head and tail preserved, all three models make correct predictions; nonetheless, including one more patch describing the bird feeder in the masked image 2 significantly alters the predictions by fine-tuning and linear probing, despite having no influence on that by the from-scratch model. This can be explained by the spurious correlations between head and bird feeder features that exist in another bird class of the pre-training dataset.

% \ying{Another source of inductive bias that is incurred by the pre-training dataset and destructive to fine-tuning is \emph{spuriously correlated features} -- misleading correlations exist in the pre-training dataset, \eg, the patch marked in the red box in Fig.~\ref{fig_3} is likely linked to another class of objects so that they mislead the final prediction of ``bird''. This accounts for why including this patch does not have any influence on the prediction by the model trained from scratch but meantime significantly alters the prediction by fine-tuning and linear probing.}

This work sets out to develop a fine-tuning strategy that alleviates the negative effects of these two types of underperforming pre-trained features.
First, we conclude that maximizing the mutual information between examples in the same class with respect to a particular slice of rare features 
is contributory to draw the attention to refining rare features to be discriminative.
% resolve these two types of biases caused by pre-trained models during fine-tuning. 
Second, by investigating the %structural 
causal models of predictions after fine-tuning, we identify that
%(1) loose coupling between discriminative and rare features, and (2) 
the pre-training dataset as a confounder explains the negative transfer by %by rare features and
spuriously correlated features. %, respectively. 
To this end, we propose to %mitigate negative transfer by (1) maximizing the mutual information between rare features and labels, and (2) 
deconfound by the principled front-door adjustment rule.
In light of the key role of a patch implementing a slice of features and the fact that each patch usually encrypts a concept, 
% By investigating the generative models of data under the first type of bias, we conclude that minimizing the conditional mutual information between informative concepts  and rare concepts given a specific label is contributory to draw the attention to refine rare features particularly.  Through a causal perspective, we identify that the pre-training dataset as a confounder explains spuriously correlated features. Grounded in the deconfounding approach of front-door adjustment, we propose to predict with both channel-level and concept-level attention; additionally, we guard the deconfounding accuracy by limiting the information contained in attentive features. In light of the key role of concepts (\ie, patches in our implementation) in debiasing both types of biases, 
we dub the proposed approach ``Concept-Tuning''.

\ying{We summarize our contributions %of this work 
as follows.
\begin{itemize}[noitemsep]
    \item We have identified two specific types of underperforming pre-trained features that give rise to negative transfer and %constructed their generative models, 
revealed the root cause of their negative impacts,
upon which more principled fine-tuning \yunqiao{techniques} can be developed.
    \item To our best knowledge, {Concept-Tuning} is the first that fine-tunes concept-level sliced features, theoretically analyzed to offer high invariance to the confounder of the pre-training dataset.
    \item {Concept-Tuning} improves the state-of-the-art sample-level fine-tuning methods by a large margin (up to $4.76\%$) and with consistency in eight \final{classification} datasets, \final{seven} pre-trained models (supervised and self-supervised), three network architectures, and different sample sizes in downstream tasks. \final{Moreover, Concept-Tuning can extend well to semantic segmentation and domain generalization tasks.}
\end{itemize}}

% \begin{table*}[!htbp]
% % \vspace{-0.15in}
% \label{tab1-1}
% 	\caption{Comparison of existing methods.}
% 	\centering
% 	\resizebox{\textwidth}{!}{
% 	\begin{tabular}{lcccccccc} 
% 	\toprule
% 	Methods 	& L2SP  & REGSL  & DELTA & BSS & StochNorm  & Co-tuning & Bi-tuning & Core-tuning\\ 
% 	\hline
% 	Regularization &  weights & weights  & features & features & statistics &  predictions &  $\times$ &  $\times$ \\
% 	Contrastive learning &  $\times$ & $\times$  & $\times$ & $\times$ & $\times$  & $\times$ & $\checkmark$ &  $\checkmark$\\
% 	\bottomrule
% \end{tabular}}
% % \vspace{-0.15in}
% \end{table*}

\section{Related Work}\label{Related Work}
\paragraph{Fine-tuning}
\ying{The majority of existing fine-tuning methods} focus on better exploiting the knowledge of a pre-trained model from different perspectives. For example, L2SP~\cite{Li2018ExplicitIB} regularizes weights of the fine-tuned model to be close to those of the pre-trained model by imposing the $\ell_2$ constraint, refraining the network from forgetting useful knowledge. REGSL~\cite{li2021improved} further introduces a layer-wise parameter regularization, where the constraint strength gradually \yunqiao{reduces} from the top to bottom layers. DELTA~\cite{li2018delta} imposes constraints on generated activate feature maps instead of weights, where the constraint on each channel is weighted by the channel importance. BSS~\cite{Chen2019CatastrophicFM} penalizes smaller singular values of learned feature representations to suppress negative transfer of spectral components. StochNorm~\cite{Kou2020StochasticN} regularizes the moving statistics in batch normalization layers to mitigate over-fitting.  Co-tuning~\cite{you2020co} proposes to explore the information encoded in the last task-specific layers of a pre-trained model, which is usually disregarded, via learning the relationship between the categories of the pre-training dataset and those of a downstream dataset. Apart from these regularization methods, Bi-tuning~\cite{zhong2020bi} \final{, Core-tuning~\cite{zhang2021unleashing} and COIN~\cite{pan2023improving}} introduce supervised contrastive learning~\cite{he2020momentum, khosla2020supervised} to better leverage the label information in the target dataset with more discriminative representations as a result. \final{However, these methods use image-wise contrastive loss, which is orthogonal to our proposed concept-wise losses.} In addition to the above methods that fine-tune all the parameters, some recent methods transfer a large-scale pre-trained model~(\eg, ViT~\cite{dosovitskiy2020image}) to downstream tasks by freezing the pre-trained backbone while tuning only a small number of newly introduced trainable parameters and the classification head. For example, VPT~\cite{jia2022visual} inserts a few learnable parameters (prompts) in input spaces of the first few layers, and  SSF~\cite{lian2022scaling} introduces a small number of learnable parameters for feature shifting after each linear layer; however, neither of them targets negative transfer specifically.

\paragraph{Spurious correlation}
Most deep neural networks are typically trained to minimize the average loss on a training set, known as empirical risk minimization~\cite{vapnik1992principles}. Despite their success, they likely suffer from \textit{spurious correlations} which hinder their successful generalization, such as the contextual correlation of co-occurring objects~\cite{singh2020don}, background correlation~\cite{xiao2020noise}, variant-features correlation~\cite{arjovsky2019invariant}. The key to solving this problem is  to tell invariant feature representations apart from %and
variant ones in the training dataset, so as to enforce the model to
%representations among the dataset and then let the model only 
rely on the invariant representations only~\cite{liu2021heterogeneous, mo2021object, luo2021rectifying}. Recently, researchers also leverage the pre-trained models to address spurious correlations in a downstream dataset~\cite{kirichenko2022last,tu2020empirical}. Different from them, we focus on alleviating negative transfer from a pre-trained model with spurious correlated features to a downstream dataset.

\paragraph{Patch representation}
The majority of existing image classification studies learn the feature representation of an entire image globally, which inevitably lose position-aware information due to the average or max-pooling operation. There has been a line of literature that learns patch-level features to deal with this problem. For example,  \yunqiao{Xiao et al.}~\cite{xiao2021region} \yunqiao{proposed} an approach that learns better feature representations of regions by maximizing the similarity in convolution features of the corresponding regions between two augmented views. \yunqiao{Xie et al.}~\cite{xie2021detco} \yunqiao{imposed} the contrastive loss between a global image and its local patches to align global and local feature representations. \yunqiao{Xie et al.}~\cite{xie2021unsupervised} first detected the foreground object regions, based on which they implemented contrastive learning of these regions to specifically improve feature representations of objects. \yunqiao{How to embrace these ideas of patch-level representation learning mainly targeting object detection towards better fine-tuning of pre-trained models remains an open but intriguing question.}

%\vspace{-0.05in}
\section{Preliminary}\label{preliminary}
% \subsection{Notations}
% \label{sec:notations}
% \paragraph{Notations}
\ying{
Before proceeding to the proposed fine-tuning approach, we first introduce the notations
that we use. %will be used. % in the following.
%and 
%self-supervised fine-tuning objectives that will be used in the following sections.
Suppose that 
we are provided with a dataset $\mathcal{D}\!=\!\mathcal{D}^{tr}\!\cup\!\mathcal{D}^{te}$, where the training set $\mathcal{D}^{tr}\!=\!\{\mathbf{x}_i^{tr},y_i^{tr}\}_{i=1}^{n_i^{tr}}$ consists of $n_i^{tr}$ training examples and the testing set  $\mathcal{D}^{te}=\{\mathbf{x}_i^{te},y_i^{te}\}_{i=n_i^{tr}+1}^{n_i^{tr}+n_i^{te}}$ is made up of $n_i^{te}$ testing examples.
$\mathbf{x}_i^{tr}$ ($\mathbf{x}_i^{te}$) and $y_i^{tr}$ ($y_i^{te}$) denote the features and label of the $i$-th training (testing) example, respectively. 
Note that the labels for the testing set are used solely for evaluation purposes.
Besides, we have access to a pre-trained model $f_\theta(\cdot)$ pre-trained on a dataset $\mathcal{D}^p$ as an initialization feature extractor.
In this paper, we aim to learn a classifier $W(\cdot)$ and fine-tune the feature extractor $f_\theta(\cdot)$ through the training set $\mathcal{D}^{tr}$, so that the function $W(f_\theta(\mathbf{x}))\mapsto y$ predicts well on the testing set $\mathcal{D}^{te}$.
}
%\vspace{-0.08in}
% \paragraph{Self-supervised fine-tuning baselines} 
\paragraph{Supervised contrastive fine-tuning}
Besides the common cross entropy loss $\mathcal{L}_{CE}(W(f_\theta(\mathbf{x})),y)$ adopted in vanilla fine-tuning, recent state-of-the-art fine-tuning methods~\cite{zhang2021unleashing,zhong2020bi} 
%In addition to the common cross-entropy loss for the predictions, both Bi-tuning~\cite{zhong2020bi} and Core-tuning~\cite{zhang2021unleashing} 
introduce the supervised contrastive loss~\cite{khosla2020supervised} for better tuning of feature representations, \ie,
\begin{align} 
 	\mathcal{L}_{con} = - \frac{1}{n} \sum_{i=1}^{n} \frac{1}{|\mathcal{P}_i|} \sum_{k^{+} \in \mathcal{P}_i} \log \frac{\exp(q_i \cdot k^{+}/ \tau) }{\sum_{k^- \in \mathcal{A}_i} \exp(q_i \cdot k^- / \tau) } . \nonumber
\label{equation：scl_loss}
\end{align}
%for the feature representations during fine-tuning. 
% Unlike the unsupervised instance discrimination approach~\cite{van2018representation} that pulls representations from two augmented views of the same image and pushes representations from different images, supervised contrastive learning takes into account the label information. 
Here $\mathcal{P}_i$ and $\mathcal{A}_i$ denote the set of positive representations from the same class to pull towards the query representation $q_i$ as an anchor, and the set of negative representations from different classes to push away, respectively; \final{$k^{+}$ and $k^{-}$ denote a sample from the positive representation set $\mathcal{P}_i$ and the negative representation set $\mathcal{A}_i$, respectively;}
% and the full set of all keys, respectively given a query representation $q_i$ as an anchor 
% Specifically, given a query representation $q_i$ as \yunqiao{an} anchor, supervised contrastive loss regards representations from the same class as the positive keys that the query should match and those from different classes as negative keys. The supervised contrastive loss is defined as:
% Here $P_i$ and $A_i$ denote the set of positive keys and the full set of all keys, respectively; 
$\tau$ is a temperature hyper-parameter.
Specifically, the representation space for application of this supervised contrastive loss could be either $q_i=W(f_\theta(\mathbf{x}^{tr}_i))$ after the classifier~\cite{zhong2020bi}  or $q_i=f_\theta(\mathbf{x}^{tr}_i)$ after the feature extractor~\cite{zhang2021unleashing}.
% The main difference between Bi-tuning and Core-tuning lies in where to apply the contrastive loss and how to generate keys. Specifically, Bi-tuning applies supervised contrastive learning on the extracted features via the backbone and the predictive values of the classification head. 
% Meanwhile, it adopts a momentum-updated network~\cite{he2020momentum} to generate keys. On the other hand, Core-tuning only applies the supervised contrastive loss on the extracted features and generates the keys from the same training batch like~\cite{chen2020simple}. Moreover, it utilizes a mixup strategy~\cite{zhang2020does} to generate hard positive/negative keys to reduce the effect of  easy-to-contrast pairs~\cite{harwood2017smart}. 
In our work, we also follow the best practice by including $\mathcal{L}_{con}$ and taking $q_i=W(f_\theta(\mathbf{x}^{tr}_i))$, which constitutes the first part of our training objective in Fig.~\ref{fig:overview}, 
% {In summary, the loss function $\mathcal{L}_{f}$ of conventional image level fine-tuning is 
\begin{equation}
    \mathcal{L}_{f} = \mathcal{L}_{CE}(W(f_\theta(\mathbf{x})),y) + \mathcal{L}_{con}.
    \label{eqn:Lf}
\end{equation}

\section{Proposed Approach}\label{methods}

\begin{figure}[t]
  \vspace{-0.1in}
	\centering
	\includegraphics[width=0.8\columnwidth]{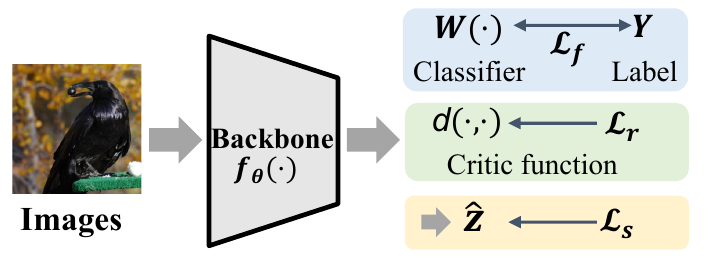}
	\caption{Overview of our proposed approach.}
	\label{fig:overview}
  \vspace{-0.1in}
\end{figure}

% \ying{We have illustrated two %sources of biases
% types of underperforming features
% in a pre-trained model that lead fine-tuning astray in Section~\ref{Introduction}, \ie, rare features and spuriously correlated features. % that are present in the pre-training dataset. 
% To mitigate such negative transfer, %the impact of 
% %by
% %these two types of biases, 
% we are motivated to first investigate their roles in 
% impairing the downstream predictions in the next, based on which we propose our principled approach.
% %the generative model of predictions 
% %during fine-tuning. 
% % % % Though the pre-training dataset $\mathcal{D}^p$ is hidden, the features $F$ that characterize an example $\mathbf{x}$  and its label $Y$ are observable variables. 
% % % We will first detail the generative models for rare features and spuriously correlated features, respectively.
% % And then based on the two types of biases in the generative models, we propose our method to alleviate the two biases. 
% Fig.~\ref{fig:overview} shows the overview, where $\mathcal{L}_r$ improves the learning of rare features and $\mathcal{L}_s$ lessens the influence of spuriously correlated features.
% }

In Section~\ref{Introduction}, we illustrated two types of underperforming features in a pre-trained model that lead fine-tuning astray: rare features and spuriously correlated features. To mitigate such negative transfer, we first investigate the roles of these features in impairing downstream predictions. Based on this investigation, we propose our principled approach. 
As shown in Fig.~\ref{fig:overview}, our approach includes $\mathcal{L}_r$, which improves the learning of rare features, and $\mathcal{L}_s$, which lessens the influence of spuriously correlated features.

\paragraph{Rare features.} 
Given a training example $\mathbf{x}^{tr}_i$, we denote its feature representation by the pre-trained model as $F_i=f_\theta(\mathbf{x}^{tr}_i)$. According to what we have established in Section~\ref{Introduction}, a rare feature $F^r_i$ is a slice of $F_i$ that should play a greater role than the other slices in classification while being not discriminative, \ie, $p(y^{tr}_i|F^r_i)\approx p(y'|F^r_i)$. $y'\neq y^{tr}_i$ here denotes the label of a different category from $y^{tr}_i$.
Imposing the example-level objective $\mathcal{L}_f$ only without focusing on slices of rare features, similar to vanilla fine-tuning, gets stuck again with those well-trained but nonessential features.

To improve the discriminative ability of rare features pointedly, we propose to maximize the mutual information between examples in the same class with respect to any $r$-th slice of rare features (\eg, neck features in Section~\ref{Introduction}), \ie,  
\begin{align}
     \max ~ I(F_i^r;F_j^r), \quad s.t. \quad y_i^{tr}=y_j^{tr}, %- I(F_R^y; F_R^{y'}), 
\label{eqn:mi_rare}
\end{align}
where $I(\cdot; \cdot)$ denotes the mutual information.
% \ie ,   maximize $I(F_{1}^{y}; F_{2}^{y}) - I(F_{1}^{y'}; F_{2}^{y})$. \textbf{[TODO: verify the correctness of the loss.]}
By zooming into each slice feature of an example, this mutual information favourably prioritizes rare features and makes them more discriminative as expected.
Despite the notorious challenge of estimating the mutual information, connections have been made of the contrastive loss to maximization of mutual information in~\cite{van2018representation,tian2020contrastive}, showing that
\begin{align}
     I(F_i^r;F_j^r) \geq \log K^- - \mathcal{L}_r,
\label{eqn:mi_rare}
\end{align}
where $K^-$ is the number of negative examples from other classes. Instead, we resort to minimize the contrastive loss 
\begin{align}
    \mathcal{L}_r = -\mathbb{E}_{\{F^r_{i}\}_{i=1}^{K^+},\{F^r_{j}\}_{j=1}^{K^+},\{F^r_{k}\}_{k=1}^{K^-}} [\log\frac{d(F^r_{i},F^r_{j})}{\sum_{k=1}^{K^-} d(F^r_{i},F^r_{k})}], \nonumber
\end{align}
where $K^+$ is the number of positive examples in the same class as in supervised contrastive loss~\cite{khosla2020supervised} and $y_i^{tr}\neq y_k^{tr}$. 

The challenge now reduces to defining the critic function $d(\cdot, \cdot)$ which is expected to evaluate the similarity between the same $r$-th slice of rare features between different examples. Take the neck features in Section~\ref{Introduction} as an example again. We seek for $d(F^r_{i}, F^r_{j})$ that pulls the neck features of the $i$-th image (characterizing $F^r_{i}$) and those of the $j$-th (describing $F^r_{j}$) given a bird class, say Green Violetear. 
Fortunately, the earth mover's distance (EMD)~\cite{andoni2008earth} function that automatically searches the most similar slice across examples meets this requirement. Therefore, we are inspired to implement 
\begin{align}
    d(F^r_{i}, F^r_{j}) = \exp(\frac{F^r_{i}\cdot \xi^*(F^{r}_{i})}{\Vert F^r_{i}\Vert\cdot \Vert \xi^*(F^{r}_{i})\Vert}\cdot \frac{1}{\tau}),
\end{align}
% By maximizing the first term $I(F_R^y;F_I^y)$, the feature $F_R^y$ will provide information similar to the informative feature $F_I^y$. Moreover, by minimizing $I(F_R^y; F_R^{y'})$, $F_R$ will be discriminative for the true class and other classes and can be used for classification.
where $\xi^* \!=\! \arg\min_{\xi:\{F_{i}^{r'}\}_{r'=1}^{R}\rightarrow \{F^{r'}_{j}\}_{r'=1}^R}\sum_{r'}\Vert F_{i}^{r'} \!-\! \xi(F_{i}^{r'}) \Vert$ represents the optimal matching flow among all one-to-one mappings $\xi\in\mathbb{R}^{R\times R}$.
$\xi^*(F_{i}^{r})$ hopefully returns the slice of features from the $j$-th example in the same semantic slice, e.g., neck features, as the $i$-th example.
Consequently, we have the empirical contrastive loss used as,
\small
\begin{align}
\mathcal{L}_r = - \frac{1}{n} \sum_{i=1}^{n} \frac{1}{K^+} \sum_{j\in\{y^{tr}_i=y^{tr}_j\}}^{K+}  \log\frac{d(F^r_{i},F^r_{j})}{\sum_{k\in\{y^{tr}_i\neq y^{tr}_k\}}^{K^-} d(F^r_{i},F^r_{k})}. 
\label{eqn:rare_features}
\end{align}
\normalsize

We leave the details of how to define a patch feature as a slice of features $F^r_i$ as well as more details on the implementation of this contrastive loss in Section~\ref{sec:exp_setup}.

\begin{figure}[t]
  \centering
  % \begin{subfigure}[t]{0.32\textwidth}
  % \centering
  % 	\includegraphics[scale=0.25]{Figs/rare_causal.pdf}%}
  % 	\caption{Rare features}
  %   \label{fig:rare_features}
  % \end{subfigure}
  % \hfill
    \begin{subfigure}[t]{0.52\columnwidth}
  \centering
  	\includegraphics[scale=0.25]{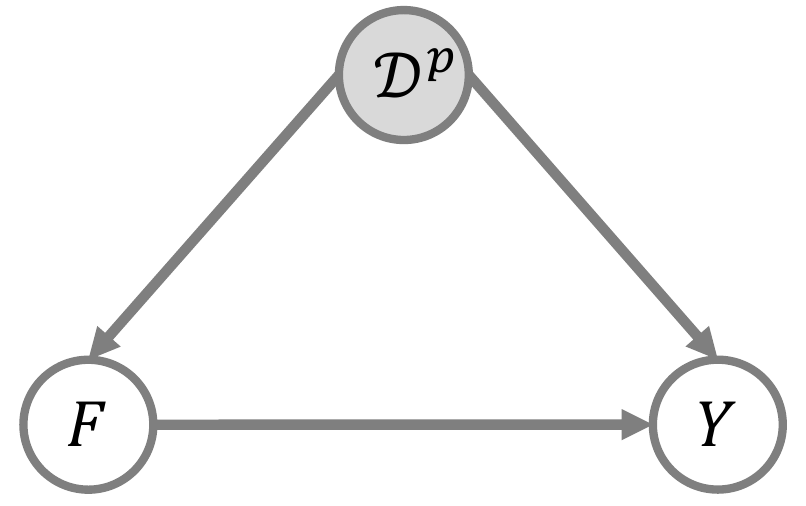}%}
  	\caption{Structural causal model of predictions}
    \label{fig:spurious_correlated_features}
  \end{subfigure}
    \hfill
      \begin{subfigure}[t]{0.47\columnwidth}
  \centering
  	\includegraphics[scale=0.25]{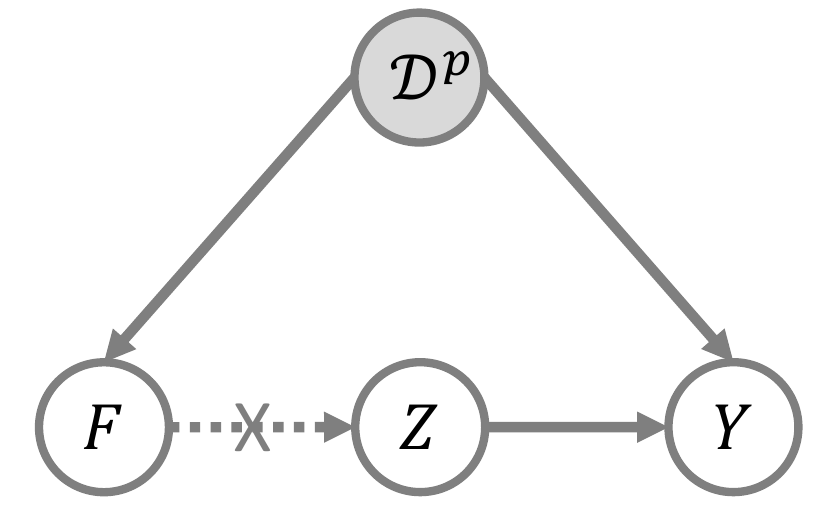}%}
  	\caption{Front-door adjustment}
    \label{fig:deconfounded_causal}
  \end{subfigure}
  \caption{A causal probe into the negative impact of spuriously correlated features in pre-trained models.}
  \label{fig:casual_probe}
  \vspace{-0.2in}
\end{figure}
\vspace{-0.1in}
\paragraph{Spuriously correlated features} 
% Suppose that $F$ can be broken down into the informative patches $F_1$ (e.g., describing a carriage) and
Despite the ground-truth label $y^{tr}_i$ (\eg, carriage),
the spuriously correlated features (\eg, describing a person) are correlated with another label $y'\neq y^{tr}_i$ (\eg, horse) more frequently in the pre-training dataset $\mathcal{D}^p$, thereby likely misguiding the prediction of $y'$, \ie,  
$
    p(y^{tr}_i|F) \ll p(y'|F)
$ as demonstrated in Section~\ref{Introduction}.
To investigate more closely the misleading role of the spuriously correlated features, we construct the structural causal model as shown
% As demonstrated in Section~\ref{Introduction}, the spurious correlations in the pre-training dataset would impose spurious correlations between the features $F$ and the label $Y$. 
%We show the causal graph 
in Fig. \ref{fig:spurious_correlated_features}. % , 
The causal link
%illustrating the pre-trained dataset $\mathcal{D}^p$ has two causal effects on the predictions. For 
%causal link 
$\mathcal{D}^p \rightarrow F \rightarrow Y$ %, $\mathcal{D}^p \rightarrow F$ 
denotes that the features $F$ extracted by the pre-trained model on the pre-training dataset $\mathcal{D}^p$ %while $F \rightarrow Y$ denotes that 
will be fine-tuned to predict
the label $Y$ via the classifier. %is predicted by using the feature $F$ with the classifier. 
% And for the causal link $\mathcal{D}^p \rightarrow Y$ indicates that the pre-trained model provides knowledge for extracting useful features for the prediction. 
% The pre-training dataset $\mathcal{D}^p$ introduces a backdoor path between $F$ and $Y$ and becomes the confounder. 
% Opposed to the existence of rare features $F_2$ in Figure~\ref{fig:rare_features}, 
% all features $F$ 
% %the spuriously correlated features $F_2$ 
% in Figure~\ref{fig:spurious_correlated_features} are assumed to be discriminative and have
% a causal effect on the final prediction $Y$, giving the causal link $F\rightarrow Y$.
Unfortunately, besides this legitimate causal path that the model is expected to learn, the ``back-door'' path $F\leftarrow \mathcal{D}^p\rightarrow Y$ gives rise to a spurious correlation from $F$ to $Y$, making $\mathcal{D}^p$ unfavorably a confounder~\cite{pearl2018book}. 
% Suppose that $F$ can be broken down into the informative patches $F_1$ (e.g., describing a carriage) and spuriously correlated features $F_2$ (e.g., describing a person). 
% Despite the ground-truth label $y$ (e.g., carriage), $F_2$ is correlated with another label $y'$ (e.g., horse) more frequently in the pre-training dataset 
% $\mathcal{D}^p$, thereby likely misguiding the prediction of $y'$, i.e.,  
% $
%     p(y|F) \ll p(y'|F).
% $
% To improve the generalization of the fine-tuned model, we should confound it. In this paper, we adopt 
% 

To this end, we propose to deconfound and thereby alleviate the negative impact of spuriously correlated features. 
% There have been two %causal inference
% deconfounding
% techniques~\cite{pearl2000models,pearl2018book}, % that %overcome the bias caused by confounders~\cite{pearl2000models,pearl2018book}, 
% including back-door adjustment that disconnects the causal link from $\mathcal{D}^p$ to $F$ and 
% front-door adjustment 
There are two deconfounding techniques~\cite{pearl2000models,pearl2018book}: back-door adjustment, which disconnects the causal link from $\mathcal{D}^p$ to $F$, and front-door adjustment, 
which introduces a mediator $Z$ from $F$ to $Y$ while severing the link from $F$ to $Z$~\footnote{Please refer to Appendix B for introduction of causal inference and deconfounding techniques.}. %front-door adjustment.}
% Note that we have provided more background materials on the front-door and back-door adjustments in Appendix B.
We adopt the front-door adjustment approach since back-door adjustment requires access to the pre-training dataset $\mathcal{D}^p$, which is always unavailable in our problem.
Concretely, as shown in Fig.~\ref{fig:deconfounded_causal}, the front-door~adjustment estimates the true causal effect between $F$ and $Y$ by deconfounding the confounder $\mathcal{D}^p$, \ie,
% The mediator $Z$ groups the features into more fine-grained clusters and provides a better representation than $F$. With the mediator, we perform the front-door adjustment as:
\vspace{-0.05in}
\resizebox{\columnwidth}{!}{
\begin{minipage}{\columnwidth}
% \begin{equation}
\small 
\begin{align}
    & p(Y|do(F)) = \sum_{z} p(Z=z|F) p(Y|Z=z)) \nonumber \\
    & = \sum_{z} p(Z=z|F) \sum_f p(F=f)[p(Y|F=f, Z=z)] \nonumber \\
    &=\mathbb{E}_{Z|F}\mathbb{E}_{F}[p(Y|F=f, Z=z)]  \nonumber \\
    & =\mathbb{E}_{Z|F}\mathbb{E}_{F}~\text{Softmax}(\varphi(F, Z)) \label{eqn:outer_expectation} \\
    & \approx \text{Softmax}[\varphi (\mathbb{E}_{F}F, \mathbb{E}_{Z|F}Z))]  \nonumber \\
    & = \text{Softmax} \big[\varphi \big( \sum_f p(F=f)\mathbf{f}, \sum_z p(Z=z|F) \mathbf{z}\big)\big], \label{eqn:front-door}
    \vspace{-0.1in}
\end{align}
% \end{equation}
\end{minipage}
}
where $do(\cdot)$ is the do-operation in causal inference, $z$ represents a slice of features from $F$. %is one of the fine-grained clusters of $F$, 
$f$ is a stratum of $F$ and $\varphi$ is a classifier.
$\mathbf{z}$ and $\mathbf{f}$ are specific embedding vectors that correspond to the two variables
$z$ and $f$, respectively.
The fourth equation comes from the implementation of $p(Y|F=f, Z=z)$ as the classifier $\varphi$ with a Softmax layer. The approximation is derived from the normalized weighted geometric mean (NWGM)~\cite{xu2015show}, which moves the outer expectation in Eqn.~(\ref{eqn:outer_expectation}) inside the Softmax function and the classifier $\varphi$. %=\varphi_1\circ \varphi_2$. 
More details on the derivation of Eqn.~(\ref{eqn:front-door}) can be found in Appendix \final{A.1}.

To be compatible with the slice-level break-down of $F$ into patches when dealing with rare features, here we formulate $Z$ as sets of patches (or concepts) specifically. 
As a result, the expectations $\mathbb{E}_F F=\sum_f p(F=f|F)\mathbf{f}$ and $\mathbb{E}_{Z|F} Z=\sum_z p(Z=z|F)\mathbf{z}$ in Eqn.~(\ref{eqn:front-door}) can be easily actualized with attention neural networks~\cite{vaswani2017attention,fu2019dual} in the level of channels and patches, respectively. We leave more implementation details of the attentional neural networks in Section~\ref{sec:exp_setup}.
% In the implementation, for the feature $F \in \mathbbm{R}^{C\times H \times W}$, where $C$ denotes the number of channels, $H$ and $W$ denote the height and width of the feature map, we extract $Z$ from $F$ using a $1\times 1$ convolution layer and obtain $z$ as the vector in each position of the feature map in $Z$. $Z$ has the same size as $F$ and we obtain $z \in \mathbbm{R}^{C}$. And we implement $\sum_z p(Z=z|F) \mathbf{z}$ by attention on $z$. We choose $f \in \mathbbm{R}^{HW}$ as the vectorized feature map in each channel of $F$ and implement $\sum_f p(F=f| F)\mathbf{f}$ by attention on $f$.

The approximation error in Eqn.~(\ref{eqn:front-door}) makes the causal link from $F$ to $Z$ not fully disconnected, thereby leaving the confounder $\mathcal{D}^p$ not fully deconfounded. 
On this account, we propose to further minimize the correlation between $F$ and the feature input $\hat{Z}$ which is measured by % to a linear classifier 
%by minimizing 
their mutual information $I(\hat{Z};F)$. 
Inspired by variational information bottleneck~\cite{alemi2017deep} where minimizing the mutual information reduces to minimizing the Kullback-Leibler (KL) divergence between
$p(\hat{Z}|F)$ and a standard Gaussian distribution $\mathcal{N}(0,1)$, we model $p(\hat{Z}|F)=\mathcal{N}(\mu^{\hat{Z}},\sigma^{\hat{Z}})$ and minimize the KL loss $\text{KL}[\mathcal{N}(\mu^{\hat{Z}},\sigma^{\hat{Z}}), \mathcal{N}(0,1)]$ instead.
Specifically, we generate the mean $\mu^{\hat{Z}}$ and variance $\sigma^{\hat{Z}}$ by projecting the concatenation of 
$\sum_f p(F=f)\mathbf{f}$ and $\sum_z p(Z=z|F)\mathbf{z}$. 
% Inspired by variational information bottleneck~\cite{alemi2017deep}, we minimize $I(\hat{Z};F)$ by minimize the KL divergence between the Gaussian distribution of $\hat{Z}$, i.e. $\mathcal{N}(\mu^{\hat{Z}},\sigma^{\hat{Z}})$ and a standard Gaussian distribution $\mathcal{N}(0,1)$ as
% $\hat{Z}=do(Z)=\varphi_2(\mathbb{E}_{F}F, \mathbb{E}_{Z|F}Z)$, which is measured by the mutual information $I(\hat{Z};F)$. 
% Inspired by variational information bottleneck~\cite{alemi2017deep} where minimizing the mutual information reduces to minimizing the Kullback-Leibler (KL) divergence between $p(\hat{Z}|F)$ and a normal Gaussian distribution $\mathcal{N}(0,1)$, we additionally predict the variance of $\hat{Z}$ via $\varphi_2(\mathbb{E}_{F}F, \mathbb{E}_{Z|F}Z)$, and model $p(\hat{Z}|F)=\mathcal{N}(\mu^{\hat{Z}},\sigma^{\hat{Z}})$ where $\mu^{\hat{Z}},\sigma^{\hat{Z}}=\varphi_2(\mathbb{E}_{F}F, \mathbb{E}_{Z|F}Z)$. 
% The resulting KL loss function that equivalently minimizes the mutual information $I(\hat{Z};F)$ is,
% \vspace{-0.05in}
% \begin{align}
%     \mathcal{L}_{kl} = \text{KL}[\mathcal{N}(\mu^{\hat{Z}},\sigma^{\hat{Z}}), \mathcal{N}(0,1)].
% \label{eqn:kl}
% \vspace{-0.05in}
% \end{align}

% Note that the classifier $\varphi$ in Eqn.~\eqref{eqn:front-door} is the composition of the mapping function $\varphi_2$ and a linear classifier $\varphi_1$.
Combining the deconfounded prediction $p(Y|do(F))$ and the minimization of $I(\hat{Z};F)$, we conclude the following loss that mitigates the spurious correlations caused by $\mathcal{D}^p$ as, 
%we have the following fine-tune 
%the classifier $\varphi$ and the attention network 
% the model by minimizing the following cross-entropy loss,
\begin{align}
    \mathcal{L}_{s} = \mathcal{L}_{CE}(p(Y|do(F)), y) + \text{KL}[\mathcal{N}(\mu^{\hat{Z}},\sigma^{\hat{Z}}), \mathcal{N}(0,1)].
\label{eqn:loss_spurious}
\end{align}

\begin{proposition}
Given that $\hat{Z}$ depends on $\mathcal{D}^p$ only through $F$, \ie, $\mathcal{D}^p\rightarrow F \rightarrow \hat{Z}$, we have
\begin{align}
    I(\hat{Z};\mathcal{D}^p) \leq I(\hat{Z};F) - I(F;Y)
\end{align}
and minimizing $I(\hat{Z};F)$ leads to invariant representation $\hat{Z}$ that is maximally insensitive to the confounder $\mathcal{D}^p$.
\label{prop:prop_1}
\end{proposition}
This proposition justifies that retaining the minimum information $I(\hat{Z};F)$ through either deconfounding or minimizing $\text{KL}[\mathcal{N}(\mu^{\hat{Z}},\sigma^{\hat{Z}}), \mathcal{N}(0,1)]$ suffices to offset the negative effects of the spuriously correlated features during fine-tuning.
%the pre-training dataset $\mathcal{D}^p$.
We provide the proof for Proposition~\ref{prop:prop_1} in Appendix \final{A.2}.

Taking all the loss functions that we have introduced for supervised contrastive fine-tuning in Eqn.~(\ref{eqn:Lf}),
% debiasing 
improving
rare features in Eqn.~(\ref{eqn:rare_features}), and alleviating spuriously correlated features in Eqn.~(\ref{eqn:loss_spurious}), we finally obtain the overall objective function that achieves concept-level fine-tuning as follows,
% \vspace{-0.1in}
\begin{align}
    \mathcal{L} = \mathcal{L}_{f} + \alpha  \mathcal{L}_{r} + \beta  \mathcal{L}_{s}
    % \vspace{-0.1in}
\end{align}
where $\alpha$ and $\beta$ are the balancing hyper-parameters that governs the impact of $\mathcal{L}_{r}$ and  $\mathcal{L}_{s}$. The pseudo-code of the whole algorithm is provided in Appendix C.

\section{Experiments}\label{Experiment}

To evaluate the effectiveness of Concept-Tuning, we conduct extensive experiments to answer the following questions: 
\textbf{Q1:} How do our methods perform compared to state-of-the-art fine-tuning methods?
\textbf{Q2:} \yunqiao{Can} the proposed {Concept-Tuning }consistently improve the performance under different data sizes? 
\textbf{Q3:} \yunqiao{Can our methods be applied to different pre-trained models and architectures?}
\textbf{Q4:} Do our methods indeed alleviate the negative impact of the two types of underperforming pre-trained features? 

\subsection{Experiment setup}\label{sec:exp_setup}
\begin{table*}[t] 
	\caption{Top-1 accuracy (\%) on various datasets using supervised pre-trained ResNet-50.}
	\label{table:comparisons}
	\centering
	\resizebox{\textwidth}{!}{\begin{tabular}{lccccccccc} 
			\toprule
			Method			 & \textbf{CUB}      & \textbf{Cars}   & \textbf{Aircraft}   & \textbf{CIFAR10}  & \textbf{CIFAR100}   & \textbf{Vegetable}    & \textbf{ISIC}    & \textbf{Caltech101}   & Avg.  \\ 
			\toprule	
			Vanilla Fine-tuning  & $78.01 \pm 0.16$ & $87.20 \pm 0.19$ & $81.13 \pm 0.21$ & $96.33 \pm 0.12$ & $83.56 \pm 0.17$ & $88.10 \pm 0.16$ &  $86.07 \pm 0.24$ &  $92.20 \pm 0.24$ & $86.56$  \\

			L2SP    			  & $78.44 \pm 0.17$ & $86.58 \pm 0.26$ & $80.98 \pm 0.29$ & $96.29 \pm 0.23$ & $83.01 \pm 0.19$ & $88.04 \pm 0.21$ & $85.99 \pm 0.22$ &  $92.36 \pm 0.17$ & $86.46$  \\
			
			DELTA   			  & $78.63 \pm 0.18$ & $86.32 \pm 0.20$ & $80.44 \pm 0.20$ & $93.77 \pm 0.16$ & $78.98 \pm 0.24$ & $88.21 \pm 0.17$ & $86.05 \pm 0.19$ & $91.95 \pm 0.33$ & $85.54$ \\
			
			BSS     		      & $78.85 \pm 0.31$ & $87.63 \pm 0.27$ & $81.48 \pm 0.18$ & $96.35 \pm 0.31$ & $83.80 \pm 0.15$ & $88.60 \pm 0.26$ & $85.55 \pm 0.17$ & $92.51 \pm 0.15$ & $86.85$   \\
			
			Co-tuning   	      & $81.24 \pm 0.14$ & $89.53 \pm 0.09$ & $83.87 \pm 0.09$ & $96.42 \pm 0.26$ & $81.40 \pm 0.22 $ & $88.26 \pm 0.19 $ & $85.21 \pm 0.13$ & $92.76 \pm 0.18$ & $87.32$   \\
			
			REGSL  			      & $81.60 \pm 0.21$ & $88.83 \pm 0.18$ & $84.07 \pm 0.23$ & $97.16 \pm 0.17$ & $83.96 \pm 0.20 $ & $87.43 \pm 0.25$ & $86.09 \pm 0.19$ & $92.21 \pm 0.22$ & $87.67$   \\
			
			% \hline
			Bi-tuning   	      & $82.93 \pm 0.23$ & $88.47 \pm 0.11$ & $84.01 \pm 0.33$ & $96.80 \pm 0.20$ & $84.44 \pm 0.18 $ & $89.60 \pm 0.26$  & $86.23 \pm 0.28$ &  $92.67 \pm 0.08$ & $88.14$   \\

			% \hline
			Core-tuning 	      & $81.99 \pm 0.12$ & $91.68 \pm 0.16$ & $86.71 \pm 0.15$ & $97.28 \pm 0.14$ & $\textbf{86.09} \pm \textbf{0.21} $ & $88.63 \pm 0.25 $ & $85.13 \pm 0.16 $ &  $91.66 \pm 0.11$  & $88.65$  \\

            \final{COIN}	   & $82.76 \pm 0.19$ & $89.88 \pm 0.21$ & $87.88 \pm 0.28$ &  $97.43\pm 0.25$ & $85.49\pm 0.22$   &  $88.52\pm 0.15$  &  $86.61\pm 0.24$  & $92.67\pm 0.18$  &  $88.90$  \\

         \hline
         Ours 1 (+$\mathcal{L}_{r}$)   & $84.86 \pm 0.26$ & $92.66 \pm 0.28$ & $88.84 \pm 0.25$ & $\textbf{97.54} \pm \textbf{0.12}$  & $85.96 \pm 0.09$  & $89.79 \pm 0.13$ & $\textbf{86.63} \pm \textbf{0.21}$  & $92.97 \pm 0.14$  & $89.91$    \\
         Ours 2 (+$\mathcal{L}_{r}$+$\mathcal{L}_{s}$)    & $\textbf{85.02} \pm \textbf{0.21}$ & $\textbf{92.90} \pm \textbf{0.24}$ & $\textbf{89.65} \pm \textbf{0.30}$ & $97.52 \pm 0.14$  & $85.59 \pm 0.15$  & $\textbf{90.10} \pm \textbf{0.18} $ & $86.01 \pm 0.26$  & $\textbf{93.15} \pm \textbf{0.10}$  & $\textbf{89.99}$   \\
        \bottomrule
	\end{tabular}}
 % \vspace{-0.2in}
\end{table*}

\noindent\textbf{Datasets.} 
We evaluate our methods on eight image classification datasets, covering a wide range of fields: {CUB-200-2011}~\cite{wah2011caltech}, {Stanford Cars}~\cite{krause20133d}, {FGVC Aircraft}~\cite{maji2013fine}, {CIFAR10}, {CIFAR100}~\cite{krizhevsky2009learning}, {Vegetable}~\cite{Hou2017VegFru}, {ISIC}~\cite{codella2019skin} and {Caltech101}~\cite{fei2004learning}.  Among them, {ISIC} is a medical dataset that targets classifying skin lesion images into seven possible disease categories. \yunqiao{We split the datasets as per previous works~\cite{you2020co,zhong2020bi} to make fair comparison. \final{Further, we apply Concept-Tuning to semantic segmentation tasks on PASCAL VOC~\cite{everingham2015pascal} and ADE20k~\cite{zhou2019semantic}, and domain generalization task on DomainNet~\cite{peng2019moment}.} We provide more descriptions for all datasets (\eg, the statics of each dataset and the train/test split) in Appendix E.1.}

\noindent \textbf{Baselines.} We \yunqiao{compare  Concept-Tuning} with recent advanced fine-tuning methods, which can be roughly categorized \yunqiao{into} three classes: regularization-based methods, supervised contrastive methods, and parameter-efficient fine-tuning methods. {(1) Regularization-based methods} including L2SP~\cite{Li2018ExplicitIB} and REGSL~\cite{li2021improved} which regularize the weights of the fine-tuned model, DELTA~\cite{li2018delta} and BSS~\cite{Chen2019CatastrophicFM} which impose constraints on the generated feature maps, and Co-tuning~\cite{you2020co} which explores label information of the pre-trained dataset as an additional regularization on the downstream predictions. (2) {Supervised contrastive methods} including Bi-tuning~\cite{zhong2020bi}, \final {Core-tuning~\cite{zhang2021unleashing} and COIN~\cite{pan2023improving}}, all of which introduce supervised contrastive learning into fine-tuning and benefit from more discriminative feature representations by pulling features from the same class together and simultaneously pushing apart features from different classes in the embedding space. \final{Specifically, COIN~\cite{pan2023improving} fine-tunes a model with $L_{con}$ in the first several epochs and later the joint of $L_{CE}$ and $L_{con}$.} (3) Parameter-efficient fine-tuning methods including VPT~\cite{jia2022visual} and  SSF~\cite{lian2022scaling}, which freeze the pre-trained backbone while tuning only a small number of newly introduced trainable parameters and the classification head. Since the last line of methods targets large-scale pre-trained models, we include them in only the experiments involving ViT-B/16. In addition, we include vanilla fine-tuning with only the cross-entropy loss as a baseline.  

\noindent\textbf{Models.} Following~\cite{zhong2020bi}, we mainly use ResNet-50 pre-trained on ImageNet-1k in a supervised manner to evaluate different fine-tuning methods on various datasets.  In addition, we conduct experiments with ResNet-50 that are pre-trained by four other self-supervised pre-training strategies~(\ie,  MoCo-V2~\cite{chen2020improved}, SimCLR~\cite{chen2020simple}, SwAV~\cite{caron2020unsupervised} and BYOL~\cite{grill2020bootstrap}) to evaluate the effectiveness of Concept-Tuning on different pre-trained models. Moreover, we follow VPT~\cite{jia2022visual} to evaluate the compatibility of the proposed method with larger models: ResNet-101 and ViT-B/16~\cite{dosovitskiy2020image}), where  ResNet-101 is supervised pre-trained on ImageNet-1K and ViT-B/16 is pre-trained on ImageNet-21K\footnote{https://github.com/rwightman/pytorch-image-models}. 

\noindent\textbf{Implementation details.} We implement our methods based on the Transfer Learning Library\footnote{https://github.com/thuml/Transfer-Learning-Library}. For the baselines, we use the same training strategies and default hyper-parameters settings as in the previous works~\cite{zhong2020bi, zhang2021unleashing}. During training, we utilize a standard augmentation strategy by performing random resize-crop to $224 \times 224$, random horizontal flip, and normalization with ImageNet means and deviations. Following MoCo~\cite{chen2020improved}, we keep a momentum-updated model and maintain two groups of queues of features. Specifically, denoting the tuning model as $\theta_{q}$ and the momentum-updated model as $\theta_{k}$, we update $\theta_{k}$ by $\theta_{k} \gets m\theta_{k} + (1-m)\theta_{q}$ where $m$ as the momentum coefficient is set as $0.999$ by default. Each queue in the first group of queues stores $n$  $\ell_2$ normalized feature representations for examples in each class and each queue in the second group stores $n$ $\ell_2$ normalized patch representations for each class, where the representations are generated by the momentum-updated model and $n$ is a hyper-parameter controlling the size of a queue. Moreover, we set the temperature $\tau = 0.07$~\cite{wu2018unsupervised}. More details of the hyper-parameter settings tuned by validation sets are provided in Appendix E.2. All results are averaged over three trials.

\noindent\textbf{Implementation details in rare features.}
To define a patch feature as a slice of features $F^r$, we first determine the regions of rare features and then extract features of the cropped regions.  In detail, we obtain the most confusing class based on the classification logits, upon which we retrieve the classification weights of the ground truth class $W_{y} \in \mathbb{R}^{C}$ ($C$ represents the number of channels in $F$) and the confusing class $W_{y'}$, respectively. 
Note that the input features $F$ to the classification head are all non-negative thanks to the ReLU operation so that $u = W_{y} - W_{y'}$  demonstrates each channel's importance in classifying these two classes. $u_i$ close to zero indicates that the $i$-th channel feature is not that discriminate. Inspired by previous works~\cite{zhou2016learning,zheng2019looking} that connect each channel of the convolutional feature map to a visual concept, we choose the attentive regions of the channels whose $u$ are closest to zero as regions of the rare features. In our experiments, we use $9$ channels with $u$ closest to zero and thereby crop $9$ patches in size of $64 \times 64$, each of which is at the most attentive position of a selected channel.

\noindent\textbf{Implementation details in spurious correlated features.}
Given a feature representation $F \in \mathbb{R}^{C \times H \times W}$ of an example, where $C$ represents the channel, $H$ and $W$ represent the height and width of the feature, we feed it into attention networks to get two expectations $\mathbb{E}_{Z|F} Z$ and $\mathbb{E}_F F$. $\mathbb{E}_{Z|F} Z$ via a patch attention module: we first obtain two new features $\{K, Q\} \in \mathbb{R}^{C/8 \times H \times W}$ to {better model the probability  $p(Z=z|F)$ in the embedded space as well as} reduce the computation complexity through two $1 \times 1$ convolution layers,  respectively. Then both $K$ and $Q$ are reshaped to $\{K_R, Q_R \} \in \mathbb{R}^{C/8 \times N}$, where $N = H \times W$ is the number of patch features.  After that, we obtain the patch-wise attention through  $P = Softmax(Q_R^T K_R) \in \mathbb{R}^{N \times N}$, where T indicates transpose operation. Meanwhile, we also obtain a new feature map $V  \in \mathbb{R}^{C \times H \times W}$ through a $1 \times 1$ convolution layer and then reshape it to $ V_R  \in \mathbb{R}^{C\times N}$. Lastly, we get the selected feature in patches via a matrix multiplication operation between $P$ and $V$: $V_{R}P \in \mathbb{R}^{C \times N}$.  $\mathbb{E}_F F$ via a channel attention module: here we directly calculate the channel attention map $S$ from $F$. Specifically, we first reshape $F$ into $F_{R} \in \mathbb{R}^{C\times N}$ and then obtain $S = F_{R}F_{R}^{T} \in \mathbb{R}^{C\times C}$ to model $p(F=f|F)$, and then  $\mathbb{E}_F F=\sum_f p(F=f|F)\mathbf{f}$  can be calculated through a matrix multiplication operation between $F_R$ and $S$: $SF_{R} \in \mathbb{R}^{C\times N}$. Finally, the two expectations are aggregated by element-wise sum operation and then average pooling along dimension $N$. Please kindly refer to  Appendix C for the pseudo-code and  Appendix D for the structure of the attention network.

\subsection{Experimental results}
\begin{table}[b] \label{tab1-1}
    \caption{Top-1 accuracy (\%) on three datasets under different sampling rates using supervised pre-trained ResNet-50.}
\label{table:data_sizes}
    \centering
    \resizebox{\columnwidth}{!}{\begin{tabular}{llccccc} 
    \toprule
    \multirow{2}{*}{Dataset} & \multirow{2}{*}{Method} & \multicolumn{4}{c}{Sampling Rates}                 \\ 
    \cline{3-6}
    &                      & 15\%       & 30\%       & 50\%       & 100\%      & Avg.   \\ 
    \hline		\multirow{5}{*}{\textbf{CUB}}  
		& Vanilla fine-tuning  & $45.25 \pm 0.12$ & $59.68 \pm 0.21$ & $70.12 \pm 0.29$ & $78.01 \pm 0.16$ & $63.27$  \\
% 		\cline{2-6}
		& Bi-Tuning   			  & $55.83 \pm 0.04$ & $69.52 \pm 0.24$ & $77.17 \pm 0.13$ & $82.93 \pm 0.23$  & $71.36$   \\
% 		\cline{2-6}
		& Core-Tuning   		  & $55.94 \pm 0.07$   & $68.54 \pm 0.16$    & $76.41 \pm 0.18$   &  $81.99 \pm 0.12$  & $70.72$   \\
        \cline{2-7}
        & Ours 1  & ${58.15} \pm 0.11$ & ${71.16} \pm 0.18$ & ${77.98} \pm 0.20$ & $84.86 \pm 0.26$   & $73.04$   \\
        & Ours 2  & $\textbf{60.70}  \pm \textbf{0.25}$ & $\textbf{73.18}  \pm \textbf{0.22}$ & $\textbf{78.68} \pm \textbf{0.14}$ & $\textbf{85.02} \pm \textbf{0.21}$  & $\textbf{74.40}$   \\

		\hline
		\multirow{5}{*}{\textbf{Cars}}  
		& Vanilla fine-tuning  & $36.77 \pm 0.12$ & $60.63 \pm 0.18$ & $75.10 \pm 0.21$ & $87.20 \pm 0.19$  & $64.93$  \\
% 		\cline{2-6}
		& Bi-Tuning   			  & $48.86 \pm 0.22$ & $73.05 \pm 0.29$ & $81.10 \pm 0.07$ & $88.47 \pm 0.11$  & $72.87$   \\
% 		\cline{2-6}
		& Core-Tuning   		  & $53.79 \pm 0.17$ & $77.27 \pm 0.24$ & $85.56 \pm 0.18$ & $91.68 \pm 0.16$  & $77.08$    \\
        \cline{2-7}

        & Ours 1  & ${55.00} \pm 0.28$ & ${78.57} \pm 0.27$  & ${86.63} \pm 0.15$  &  $92.66 \pm 0.28$  & $78.22$    \\
        & Ours 2  & $\textbf{57.32}  \pm \textbf{0.19}$ & $\textbf{79.99} \pm \textbf{0.22}$ & $\textbf{87.60} \pm \textbf{0.12}$ & $\textbf{92.90} \pm \textbf{0.24}$  & $\textbf{79.45}$   \\

		\hline
		\multirow{5}{*}{\textbf{Aircraft}} 
		& Vanilla fine-tuning  & $39.57 \pm 0.20$ & $57.46 \pm 0.12$ & $67.93 \pm 0.28$  & $81.13 \pm 0.21$  & $61.52$  \\
% 		\cline{2-6}
		& Bi-Tuning   & $47.91 \pm 0.32$ & $64.45 \pm 0.23$  & $72.40 \pm 0.22$ & $84.01 \pm 0.33$  & $67.19$   \\
% 		\cline{2-6}
		& Core-Tuning   		  & $\textbf{50.38} \pm \textbf{0.34} $ & $68.78 \pm 0.28 $ & $77.86 \pm 0.26 $  & $86.71 \pm 0.15$   & $70.93$   \\
        \cline{2-7}
        & Ours 1  & $49.02 \pm 0.28$  &  $68.05 \pm 0.17$ & $78.04 \pm 0.16$  &  $88.84 \pm 0.25$  & $70.99$  \\
        & Ours 2  & $50.05 \pm 0.23$ & $\textbf{69.10} \pm \textbf{0.27}$ & $\textbf{78.40} \pm \textbf{0.24}$ & $\textbf{89.65} \pm \textbf{0.30}$  & $\textbf{71.80}$ \\

 \bottomrule
\end{tabular}}
\end{table}

\noindent \textbf{Comparison with previous methods.}
We first report the performances of different methods on \final{eight image classification} datasets using ResNet-50~\cite{he2016deep} supervised pre-trained on ImageNet-1k in Table~\ref{table:comparisons}. Concept-Tuning outperforms all competitors by a large margin (average $1.34\%$ improvements on eight datasets), especially on fine-grained datasets (\eg, $2.09\%$ improvements on CUB and $2.94\%$ on Aircraft compared to the previous best method). Strictly regularizing the weights or features (\eg, L2SP and DELTA) between the pre-trained and fine-tuned models could exacerbate negative transfer and thus achieve poor performances, which is in most cases even worse than vanilla fine-tuning. Relaxing the weights-regularization on different layers as REGSL or only penalizing smaller singular values of features as BSS could alleviate the forehead problem and achieve better performances. Moreover, introducing supervised contrastive learning during fine-tuning(\eg, Bi-tuning and Core-tuning) can achieve superior performances as it \yunqiao{efficiently leverages} the label information via supervised contrastive learning. \final{We attribute less improvement on CIFAR to the low resolution (32×32) images, which makes accurate extraction of concept-level features challenging. However, this does not limit the applicability of Concept-Tuning in light of prevailing high-resolution images nowadays. As shown in the averaged accuracy on eight datasets, our method achieves the best performance by successfully reducing the negative transfer caused by the two underperforming features in the pre-trained model.}

\noindent\textbf{Results with different data sizes.}
To verify the performance of our methods on different sizes of training data,  especially on a small training dataset, we follow \cite{you2020co} to sample $15\%$, $30\%$, $50\%$ and $100\%$ of data for training. We show the performances of the proposed method and a part of the baselines in Table~\ref{table:data_sizes} and the full table in Appendix F. \yunqiao{We find that our method consistently surpasses all previous methods} under different sampling rates. This indicates that even when the training data is limited, \yunqiao{our methods can resolve the severe negative transfer in the pre-training model and achieve much better performance than the previous methods. For example, under $15\%$ sampling rate on CUB, our method is $4.87\%$ and $4.76\%$ higher than Bi-tuning and Core-tuning, respectively.}

\noindent\textbf{Results with different unsupervised pre-training strategies.}
To evaluate the effectiveness of our methods on unsupervised pre-trained models, we further conduct extensive experiments on ResNet-50 pre-trained by four widely-used self-supervised pre-training methods \final{MoCo-V2~\cite{chen2020improved}, SimCLR~\cite{chen2020simple}, SwAV~\cite{caron2020unsupervised} and BYOL~\cite{grill2020bootstrap} and ViT-B/16 pre-trained by two self-supervised pre-training methods (MAE~\cite{he2022masked} and MoCo-v3~\cite{chen2021empirical})}.  
All the models are pre-trained on ImageNet and the pre-trained weights are provided by their authors. As shown in Table~\ref{tabel:pre-trained-trategies}, our methods work well under different self-supervised pre-trained methods, showing that the effectiveness of Concept-Tuning is not bounded to specific pre-training models. More results \final{(\eg, MAE~\cite{he2022masked} and MoCo-v3~\cite{chen2021empirical}))} are available in Appendix F.

\begin{table}[b] 
% \vspace{-0.1in}
	\caption{Top-1 accuracy (\%) on three datasets using four different pre-trained ResNet-50.}
	\label{tabel:pre-trained-trategies}
	\centering
	\resizebox{\columnwidth}{!}{
	\begin{tabular}{llccccc}
		\toprule
		\multirow{2}{*}{Dataset} & \multirow{2}{*}{Method} & \multicolumn{4}{c}{Pre-trained method}  \\
		\cmidrule{3-6}  
		&   & MoCo-V2      &  SimCLR    & SwAV   &  BYOL  &  Avg.    \\ 
		\hline
		\multirow{5}{*}{\textbf{CUB}}    
		& Vanilla fine-tuning    & $76.72 \pm 0.21$ & $76.51 \pm 0.28$ & $80.45 \pm 0.32$ & $81.29 \pm 0.29$ & $78.74$  \\
		& Bi-tuning   			 & $79.48 \pm 0.24$ & $75.73 \pm 0.25$ & $81.72 \pm 0.23$ & $82.02 \pm 0.29$  & $79.74$\\
		& Core-tuning            & $77.93 \pm 0.18$ & $77.55 \pm 0.15$ & $80.60 \pm 0.27$ & $78.46 \pm 0.18$  & $78.64$  \\
            &  Ours 1  & ${82.48} \pm 0.14$ & ${78.18}  \pm 0.20$ & ${83.47}  \pm 0.22$ & ${83.38}  \pm 0.18$  & $81.88$     \\
            &  Ours 2 & $\textbf{82.53} \pm \textbf{0.21}$ & $\textbf{79.81} \pm \textbf{0.23}$ & $\textbf{84.78} \pm \textbf{0.32}$ & $\textbf{84.45} \pm \textbf{0.29}$  & $\textbf{82.89}$   \\
		
		\hline
		\multirow{5}{*}{\textbf{Cars}}    
		& Vanilla fine-tuning     & $88.45 \pm 0.35$ & $84.53 \pm 0.12$ & $88.17 \pm 0.21$ & $88.99 \pm 0.39$  & $87.54$ \\  
		& Bi-tuning   			  & $90.05 \pm 0.15$ & $91.75 \pm 0.18$ & $90.49 \pm 0.27$ & $90.90 \pm 0.18$  & $90.80$\\	
		& Core-tuning             & $90.87 \pm 0.23$ & $91.78 \pm 0.26$ & $91.84 \pm 0.14$ & $91.95 \pm 0.18$  & $91.61$ \\
            &  Ours 1 & ${91.02} \pm 0.11$ & ${93.27} \pm 0.20$ & ${93.41} \pm 0.26$ & ${93.22} \pm 0.15$   & $92.73$ \\
            &  Ours 2 & $\textbf{91.75} \pm \textbf{0.18}$ & $\textbf{93.36} \pm \textbf{0.23}$ & $\textbf{93.79} \pm \textbf{0.32}$ & $\textbf{93.68} \pm \textbf{0.25}$  & $\textbf{93.15}$   \\

		\hline
		
		\multirow{5}{*}{\textbf{Aircraft}}   
		& Vanilla fine-tuning   & $88.60 \pm 0.18$ & $87.79 \pm 0.24$ & $83.26 \pm 0.17$ & $85.03 \pm 0.15$  & $86.17$ \\
		& Bi-Tuning         	& $89.05 \pm 0.16$ & $88.69 \pm 0.17$ & $85.69 \pm 0.13$ & $87.16 \pm 0.11$  & $87.65$  \\
		& Core-Tuning   	 & $89.02 \pm 0.19$ & $89.47 \pm 0.21$ & $88.66 \pm 0.34$ & ${89.74} \pm {0.20}$  & $89.22$ \\
            & Ours 1  & $\textbf{89.65} \pm \textbf{0.18}$ & ${90.13} \pm 0.11$ & ${91.42} \pm 0.36$ & ${90.82} \pm 0.22$  & $90.50$ \\
            &  Ours 2  & ${89.32} \pm 0.21$ & $\textbf{90.85} \pm \textbf{0.17}$ & $\textbf{91.75} \pm \textbf{0.14}$ & $\textbf{91.21} \pm \textbf{0.13}$  & $\textbf{90.76} $   \\

		\bottomrule
		\end{tabular}
	}
 % \vspace{-0.1in}
\end{table}

\begin{table}[t] 
% \vspace{-0.2in}
	\caption{Top-1 accuracy (\%) on two datasets using   different architectures with supervised pre-training, where $*$ indicates the results reported in ~\cite{jia2022visual} and ~\cite{lian2022scaling}.}
	\label{tabel:archs}
	\centering
	\resizebox{0.9\columnwidth}{!}{
	\begin{tabular}{llcccc}
		\toprule
		\multirow{2}{*}{Dataset} & \multirow{2}{*}{Method} & \multicolumn{3}{c}{Architecture}  \\
		\cmidrule{3-5}  
		&  &    ResNet-50      & ResNet-101     & ViT-B/16 & Avg.\\ 
		\hline

		\multirow{7}{*}{\textbf{CUB}}    
		& Vanilla fine-tuning    & $78.01 \pm 0.16$ & $82.26 \pm 0.34$ & $87.3^*$  & $81.32$\\
		& Bi-tuning   			 & $82.93 \pm 0.23$ & $83.57 \pm 0.13$ & $89.23 \pm 0.33$   & $85.24$\\
		& Core-tuning            & $81.99 \pm 0.12$ & $81.76 \pm 0.14$ & $89.30 \pm 0.21$   & $84.35$\\
  		& VPT			         & * & *  &  $88.5^*$ & *\\
		& SSF                    & * & *  &  $89.5^*$ & * \\
            & Ours 1              & ${84.86} \pm 0.26$ & ${83.98} \pm 0.17$ & $\textbf{89.90} \pm \textbf{0.18}$  & $86.25$ \\
            & Ours 2              & $\textbf{85.02} \pm \textbf{0.21}$ & $\textbf{84.12} \pm \textbf{0.16}$ & $89.78 \pm 0.25$  & $\textbf{86.31}$  \\

		\hline
		\multirow{7}{*}{\textbf{Cars}}    
		& Vanilla fine-tuning     & $87.20 \pm 0.19$ & $88.84 \pm 0.28$ & $84.5^*$ & $86.85$ \\  
		& Bi-tuning   			  & $88.47 \pm 0.11$ & $90.82 \pm 0.16$ & $92.40 \pm 0.19$  & $90.56$ \\	
		& Core-tuning             & $91.68 \pm 0.16$ & $91.49 \pm 0.13$ & $91.12 \pm 0.34$  & $91.43$\\
  		& VPT			         & * & *  &  $83.6^* $ & * \\
		& SSF                    & * & *  &  $89.2^*$  & * \\
            & Ours 1              & ${92.66} \pm 0.17$ & ${92.63} \pm 0.22 $ & ${92.86} \pm 0.19$ & $92.72$   \\
            & Ours 2             & $\textbf{92.90} \pm \textbf{0.24}$ & $\textbf{92.76} \pm \textbf{0.24}$ & $\textbf{92.92} \pm \textbf{0.26}$ & $\textbf{92.86}$  \\
            
		\bottomrule
		\end{tabular}
 	}
  % \vspace{-0.1in}
\end{table} 

\noindent\textbf{Results with different backbone architectures.}
Previously we conduct experiments on ResNet-50, while the effectiveness of Concept-Tuning on more architectures is unexplored. Here, we conduct experiments on three different backbone architectures: ResNet-50, ResNet-101~\cite{he2016deep}, which are supervised pre-trained on ImageNet-1k, and Vit-B/16~\cite{dosovitskiy2020image} supervised pre-trained on ImageNet-21k. As shown in Table~\ref{tabel:archs}, Concept-Tuning performs well on all three architectures. Despite the superior performances of vanilla fine-tuning on ViT-B/16, Concept-Tuning obtains non-trivial performance gains (\eg, $2.6\%$ on CUB), verifying the effectiveness of Concept-Tuning on large pre-trained models.

\subsection{Ablation studies}
\noindent \textbf{The effect of the two losses $\mathcal{L}_r$ and $\mathcal{L}_s$.}
We first conduct ablation studies of Concept-Tuning on the loss regarding rare features $\mathcal{L}_{r}$ \final{, spuriously correlated features $\mathcal{L}_{s}$, and invariant representation KL.} As shown in Table~\ref{table:ablation},  only resolving rare features via $\mathcal{L}_{r}$ can improve the performances by a large margin, indicating that it leads to better feature representations and thus achieves better classification results, and only resolving spuriously correlated features to disconnect causal links via $\mathcal{L}_{s}$ can also obtain obvious improvements. \final{Further, minimizing $I(\hat{Z},F)$ through the proposed KL loss promotes invariant representations $\hat{Z}$, verified by the improvements on ${L}_{s}$ beyond   ${L}_{s}$ w/o KL.}  Moreover, we find that simultaneously applying $\mathcal{L}_{r}$ and $\mathcal{L}_{s}$ yields the best results, as shown in the last row of Table~\ref{table:ablation}.

\begin{table}[b] 
% \vspace{-0.1in}
	\caption{Ablation studies of our methods using ResNet-50 by supervised pre-training, and we report the top-1 accuracy (\%) on three datasets. Note that we follow the training strategies in Bi-tuning~\cite{zhong2020bi} to implement the experiments with $\mathcal{L}_{f}$, which is a stronger baseline.}
	\label{table:ablation}
	\centering
	\resizebox{\columnwidth}{!}{
	\begin{tabular}{cccc|ccc}
		\toprule
            \multicolumn{4}{c}{Loss} & \multicolumn{3}{c}{Dataset}  \\
		  $\mathcal{L}_{f}$ & $\mathcal{L}_{r}$ & $\mathcal{L}_{s}$  & \final{$\mathcal{L}_{s}$ w/o KL}  &  {CUB}  & {Cars}  & {Aircraft} \\
            \hline
  	 \checkmark  &   &  &   & $82.93 \pm 0.23$ & $88.47 \pm 0.11$ & $84.01 \pm 0.33$ \\  %% bi-tuning paper 

		\checkmark &   \checkmark   &  &    & $84.86  \pm 0.26$ & $92.66  \pm 0.28$ & $88.84  \pm 0.25$ \\
  
    	\checkmark  &   &   & \final{\checkmark}   & $ {83.36} \pm {0.25}$ & $ {90.49} \pm {0.24}$ & ${86.29} \pm {0.31}$\\

		\checkmark  &  &   \checkmark  &   & $84.21  \pm 0.23$ & $91.84  \pm 0.19$ & $87.46  \pm 0.28$ \\
  
  		\checkmark  &  \checkmark &   & \final{\checkmark}   & $ {84.92} \pm {0.22}$ & $ {92.55} \pm {0.28}$ & ${89.11} \pm {0.36}$\\
    
		\checkmark  &  \checkmark &  \checkmark    &  & $ \textbf{85.02} \pm \textbf{0.21}$ & $\textbf{92.90} \pm \textbf{0.24}$ & $\textbf{89.65} \pm \textbf{0.30}$ \\
            
		\bottomrule
		\end{tabular}
 	}
% \vspace{-0.1in}
\end{table}

More ablation studies~(\ie, the trade-off weight $\alpha$, the trade-off weight $\beta$, the patch size, the temperature $\tau$, and the number of keys stored in the queue) are provided in Appendix F.

\subsection{Visualization}
\begin{figure}[!t]
	\begin{subfigure}[]{0.4\columnwidth}
		\centering
		{\includegraphics[width = 0.9\columnwidth]{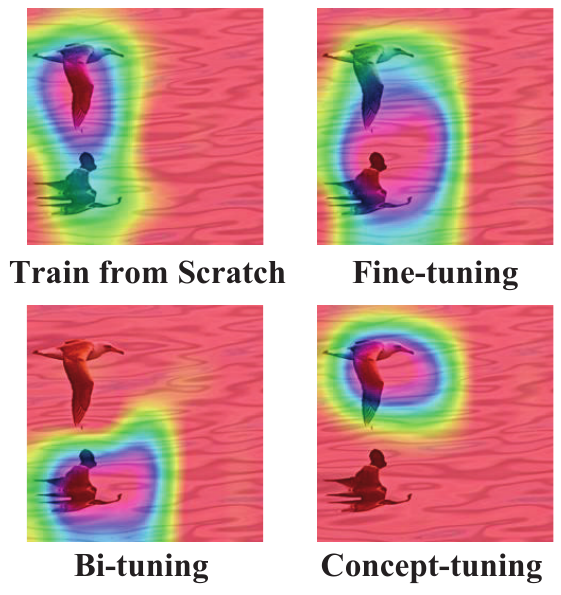}}
      	\caption{}
		\label{fig:ours_cam}
	\end{subfigure}
	\begin{subfigure}[]{0.54\columnwidth}
		\centering
		{\includegraphics[width = \columnwidth]{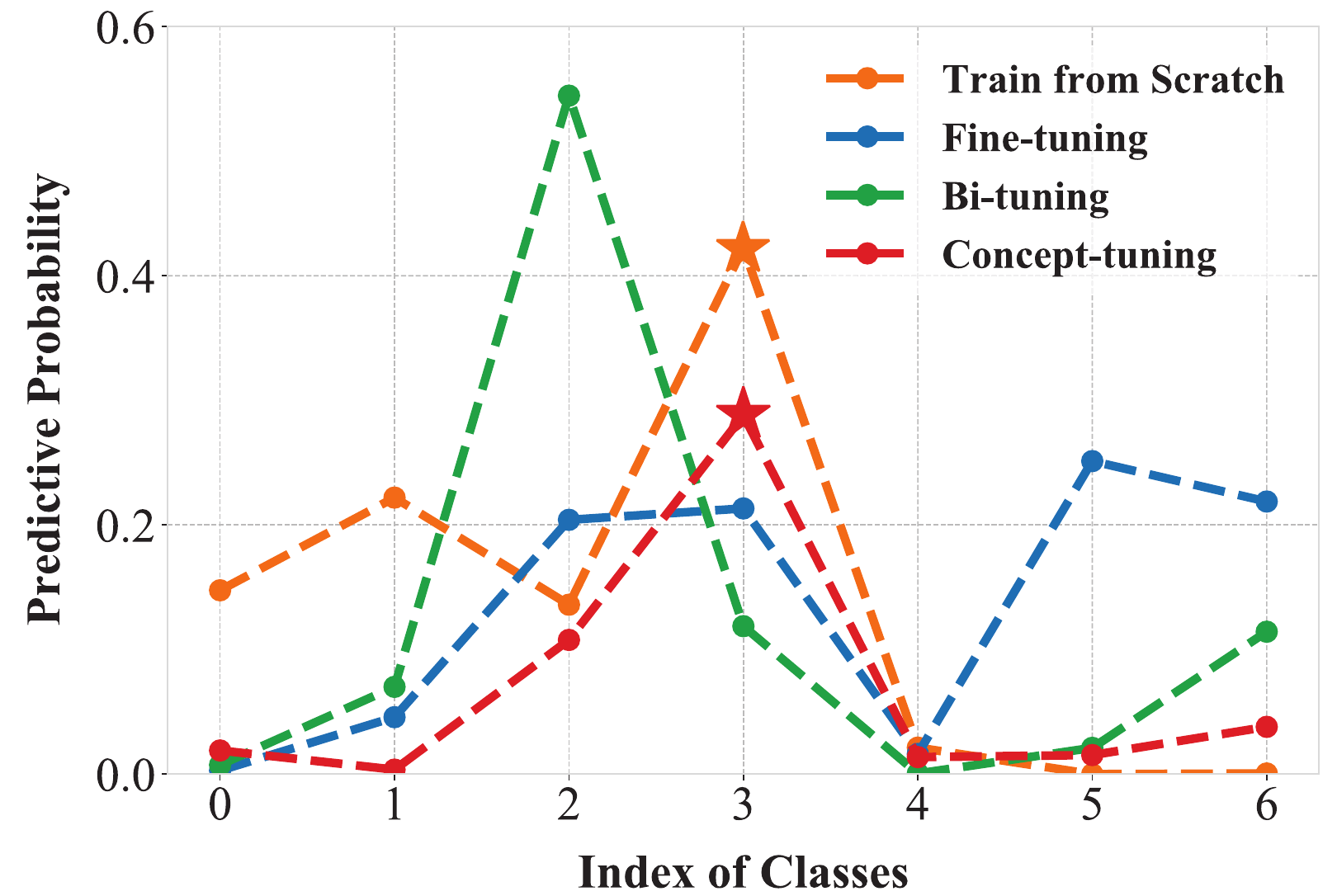}}
      	\caption{}
	       \label{fig:ours_pred_probabilities}
	\end{subfigure} 
 \caption{(a) CAM visualization of 4 methods. Influenced by the pre-trained model, fine-tuning and Bi-tuning will be attracted by the reflection in the water and make wrong predictions, while Concept-Tuning focuses on the bird and predicts correctly. (b) Given a Gaussian-blurred masked image in Fig.~\ref{fig:similar_images}, our method alleviates the negative transfer and predicts correctly as the model trained from scratch.}
% \vspace{-0.2in}
\end{figure}

To better understand the effectiveness of our methods, we compare the CAMs of different methods, as shown in Fig.~\ref{fig:ours_cam}. Influenced by the pre-trained model, Fine-tuning and Bi-tuning tend to focus on the reflection in the water. In contrast, our methods could alleviate the problem and attend to regions closer to the model trained from scratch. Moreover,  Concept-Tuning focuses on a relatively minor part than Bi-tuning, indicating that our method could effectively resolve those spuriously correlated features. To further verify that our approach can improve the representation of rare features, we test our methods on the masked image in Fig.~\ref{fig:similar_images}. As shown in Fig.~\ref{fig:ours_pred_probabilities}, Concept-Tuning predicts correctly, while both Fine-tuning and Bi-tuning fail.

\begin{figure} 
    \centering
    \includegraphics[width=0.7\columnwidth]{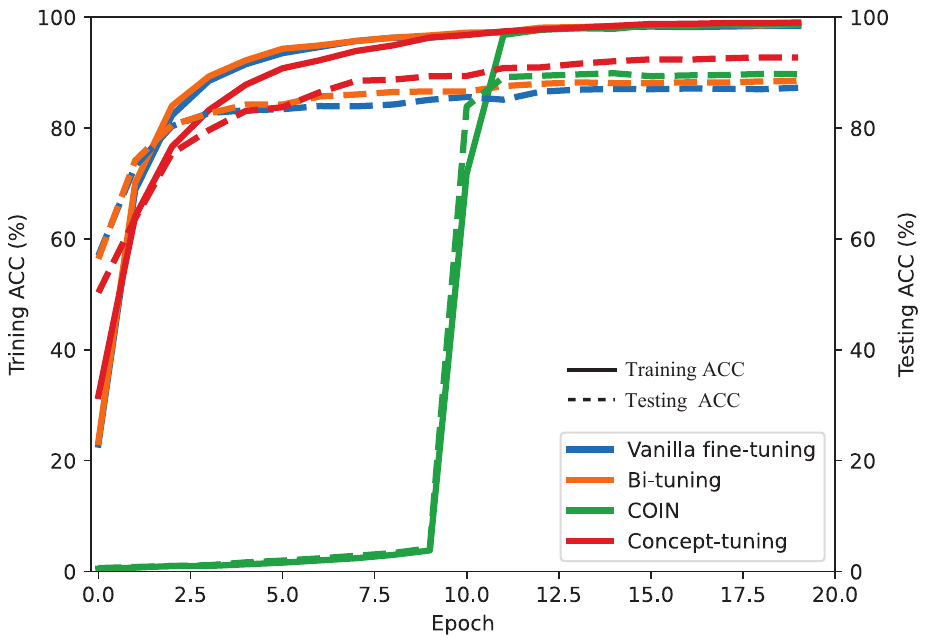}
    \caption{\final{Comparisons of the training curves of different methods on Cars.}}
    \label{fig:training_curve}
\end{figure}

\final{To compare the convergence of different methods, we follow COIN~\cite{pan2023improving} to plot the training curves. As shown in Fig.~\ref{fig:training_curve}, all methods have fully converged to similar training accuracy, while Concept-Tuning achieves the best performance on the testing set.}

\subsection{Experiments on semantic segmentation}
\final{
To explore the feasibility of our methods beyond visual classification, we extend the evaluation of ours to~semantic segmentation, where we follow the training protocol~in~\cite{mmseg2020} to fine-tune the pre-trained DeepLab-V3 model~\cite{chen2017rethinking}  (ResNet-50 as the backbone) on PASCAL VOC~\cite{everingham2015pascal} and ADE20k~\cite{zhou2019semantic}, respectively. Following~\cite{mmseg2020}, for PASCAL VOC, we train on the VOC2012 training set and its augmented training set~\cite{hariharan2011semantic} with 20k iterations and then test on VOC2012 validation set;  for ADE20k, we train on its training set with 80k iterations and test on its validation set. We use SGD  with an initial learning rate of $0.01$  and a poly decay schedule to train the model. The images are resized to $512 \times 512$ for PASCAL VOC and ADE20k, and other hyper-parameters are the same as~\cite{mmseg2020}. We use Mean Intersection over Union (MIoU) to evaluate the performances of different methods. Results in Table~\ref{table:segmentation} corroborate the effectiveness of Concept-Tuning in even segmentation tasks.}

\begin{table}[!hbt]   
    \centering
    \captionof{table}{ {Results on fine-tuning DeepLab-V3 (ResNet-50) on two semantic segmentation datasets.}}
    \label{table:segmentation}
    \resizebox{0.75\columnwidth}{!}{
	\begin{tabular}{lcc}
		\toprule
		 {Method}  & PASCAL VOC     & ADE20k      \\ 
  		\hline
            Vanilla fine-tuning    & $76.17$ & $42.42$   \\
		Core-tuning            & $76.94$ & $42.97$  \\
            \textbf{Ours 2}         & $\textbf{77.66}$ & $\textbf{43.12}$  \\
		\bottomrule
		\end{tabular}
    }
\end{table}

\begin{table}[!hbt] 
    \caption{\final{OOD generalization results on DomainNet. Moreover, c/i/p/q/r/s represent different domains; :c means that the model is fine-tuned on domains except for c and evaluated on c.}}
    \label{table:domainnet}
    \centering
    \resizebox{\columnwidth}{!}{
    \begin{tabular}{lc|cccccc}
    \toprule
      Method & avg. & :c & :i & :p  & :q & :r & :s \\
      \hline
     Vanilla fine-tuning  & 45.72  & 62.98  & 26.31 & 52.63 & 14.20  & \textbf{67.00} & 51.25 \\ 
     Bi-tuning            & 46.32  & 64.42  & 27.65 & 53.18 & 14.79  & 66.36 & 51.56 \\ 
     Core-tuning          & 46.06  & 63.68  & 26.94 & 52.89 & 15.17  & 65.91 & 51.78 \\ 
     \textbf{Ours 2}  & \textbf{46.74} &  \textbf{64.48} &  \textbf{28.17} & \textbf{53.68} &  \textbf{15.33} & {66.04}  &  \textbf{52.73}\\ 
    \bottomrule
    \end{tabular}
    }
\end{table}

\subsection{Benefits to  out-of-distribution generalization}
\final{We also conduct experiments on the domain generalization task to investigate our method's robustness on out-of-distribution (OOD) datasets. We compare methods on a large-scale multi-source dataset, \ie, DomainNet~\cite{peng2019moment}. DomainNet includes six domains (\ie, Clipart, Infograph, Painting, Quickdraw, Real, and Sketch), and the model is fine-tuned on five domains and evaluated on one remaining domain. Specifically, we use supervised pre-trained ResNet-50 as the backbone. For training, we follow the training protocol in Transfer Learning Library to set the learning rate as  $0.01$ and the training step as  $50,000$.  Results in table~\ref{table:domainnet} show that Concept-Tuning surpasses previous fine-tuning methods, advocating its advantage in generalizing to OOD datasets.}

\section{Conclusions and Discussions}\label{conclusions}
\ying{Drew on our preliminary empirical observations, we pinpoint two types of underperforming pre-trained features in pre-trained models that likely give rise to negative transfer, \ie, rare features and spuriously correlated features. In this paper, we develop a highly principled Concept-Tuning, which combats the negative impacts of rare features from an information-theoretic perspective and that of spuriously correlated features based on causal adjustment.
By formulating concepts as patches, we have concretely derived the concept-level contrastive loss function and the prediction with attention to channels and patches.
Extensive experiments validate the effectiveness of the proposed Concept-Tuning in a handful of downstream tasks with various pre-trained models and under different sample sizes. \textbf{Limitations:} we are more than eager to investigate the scope and types of downstream tasks which the proposed Concept-Tuning significantly boosts, which is only partially understood insofar.
}

% \section*{Acknowledgement}

{\small
\bibliographystyle{ieee_fullname}
\bibliography{references}
}

\newpage
\appendix
% \clearpage
\section{Proofs}

\subsection{Derivation of Eqn. (7) with NWGM approximation}
The work of~\cite{baldi2014dropout} gave the approximation to the expectation $\mathbb{E}_F [g(F)]$ as,
\begin{align}
    \mathbb{E}_F [g(F)]\approx \prod_F g(F)^{p(F)},
\end{align}
based on which we have the expectation of the softmax function $\mathbb{E}_F [\sigma(g(F))]$ as,
\begin{align}
    \mathbb{E}_F [\sigma(g(F))] & \approx \frac{\prod_F \exp(g_y(F))^{p(F)}}{\sum_{k}\prod_F \exp(g_k(F))^{p(F)}} \nonumber \\
    & = \frac{\exp({\sum_F} g_y(F)p(F))}{\sum_k\exp({\sum_F} g_k(F)p(F))} \nonumber \\
    & = \sigma(\mathbb{E}_F g(F)) 
    \label{eqn:wgm}
\end{align}
Following~Eqn.(\ref{eqn:wgm}), we derive
\begin{align}
    & p(Y|do(F)) \nonumber  \\
    =& \sum_z p(Z=z|F)\sum_f p(F=f)[p(Y|F=f, Z=z)] \nonumber \\
    =&\mathbb{E}_{Z|F}\mathbb{E}_{F} [p(Y|F,Z)] \nonumber \\
    =& \mathbb{E}_{Z|F}\mathbb{E}_{F} \sigma(\varphi{_1}(F,Z)) \nonumber \\
    \approx & \sigma[\mathbb{E}_F\mathbb{E}_{Z|F} \varphi{_1}( F,  Z)].
    \label{eqn:outer_approx}
\end{align}
Provided that the classifier $\varphi_1$ is linear with respect to either $\mathbb{E}_F F$ or $\mathbb{E}_{Z|F} Z$, we have the further approximation of Eqn.~(\ref{eqn:outer_approx}) as,
\begin{align}
    \sigma[\mathbb{E}_F\mathbb{E}_{Z|F} \varphi{_1}( F,  Z)] \approx \sigma[\varphi{_1}(\mathbb{E}_F F, \mathbb{E}_{Z|F})],
\end{align}
which completes the derivation.
\subsection{Proof of Proposition 4.1}
\begin{proof}
We have the causal chain $(\mathcal{D}^p, Y)\rightarrow F \rightarrow Z$, so that $I(\hat{Z};Y,\mathcal{D}^p)\leq I(\hat{Z};F)$.
The left part $I(\hat{Z};Y,\mathcal{D}^p)=I(\hat{Z};\mathcal{D}^p)+I(\hat{Z};Y|\mathcal{D}^p)$ according to the chain rule of mutual information.
Combining both, we have
\begin{align}
    I(\hat{Z};\mathcal{D}^p)\leq I(\hat{Z};F)-I(\hat{Z};Y|\mathcal{D}^p).
\label{eqn:prop_proof_ieq1}
\end{align}
According to the definition of mutual information, we have
\begin{align}
    I(\hat{Z};Y|\mathcal{D}^p) &= H(Y|\mathcal{D}^p) - H(Y|\hat{Z},\mathcal{D}^p) \nonumber \\
    &=H(Y)-H(Y|\hat{Z},\mathcal{D}^p) \nonumber \\
    &\geq H(Y) - H(H(Y|\hat{Z}) = I(\hat{Z};Y),
\label{eqn:prop_proof_ieq2}
\end{align}
which is conditioned on the fact that the label $Y$ is independent from the pre-training dataset $\mathcal{D}^p$.
Substituting the inequality~(\ref{eqn:prop_proof_ieq2}) into inequality~(\ref{eqn:prop_proof_ieq1}), we finally obtain
\begin{align}
    I(\hat{Z};\mathcal{D}^p)\leq I(\hat{Z};F) - I(\hat{Z};Y) = I(\hat{Z};F) - I(F;Y)
\end{align}
which holds under the assumption that $\hat{Z}$ is sufficient for $F$.
\end{proof}

\section{Background on Causal Intervention}
\begin{figure}[t]
	\centering
	\begin{subfigure}[b]{0.48\columnwidth}
		\centering
			\includegraphics[scale=0.3]{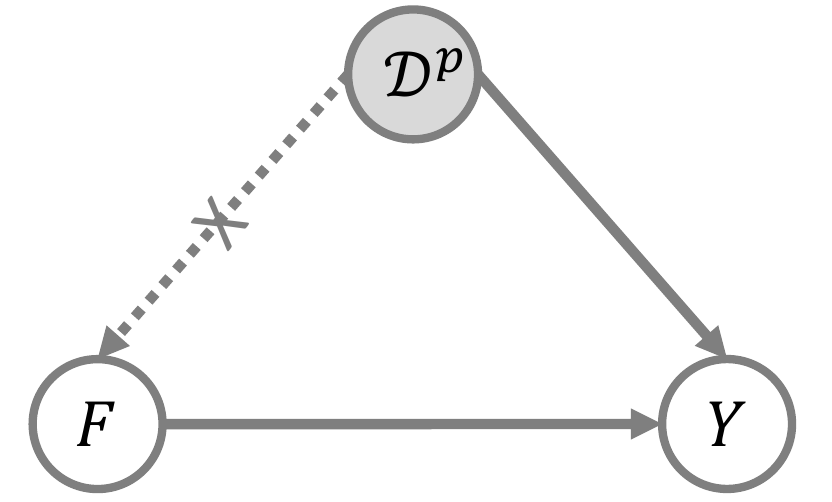}
		\caption{Backdoor adjustment}
		\label{fig:back_door}
	\end{subfigure}
	\begin{subfigure}[b]{0.48\columnwidth}
		\centering
		\includegraphics[scale=0.3]{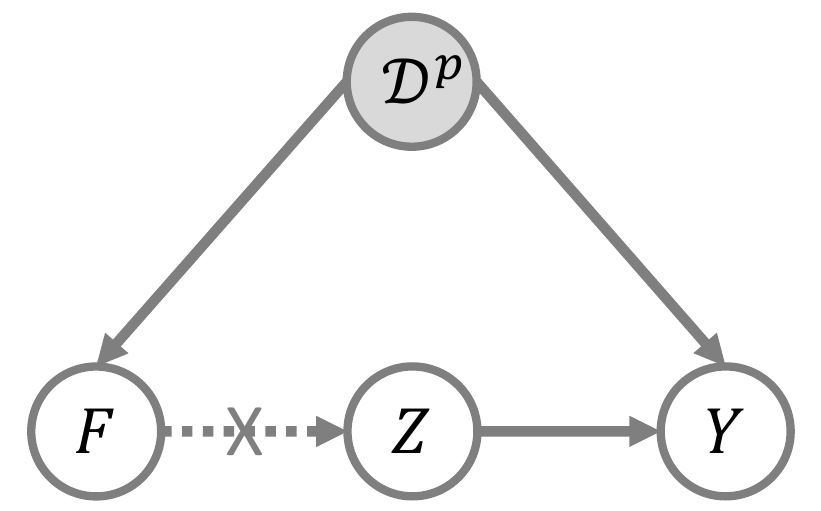}
	\caption{Front-door adjustment}
	\label{fig:front_door}
\end{subfigure}
\caption{Two categories of deconfounding methods.}
\end{figure}
\ying{The primary objective of causal intervention is to deconfound the confounder, \eg, $\mathcal{D}^p$ in our problem. There are mainly two categories of deconfounding techniques, which we will detail below.}
\subsection{Backdoor adjustment}
\ying{The backdoor adjustment seeks a way to block the causal link from the confounder $\mathcal{D}^p$ to the features $F$, as shown in Fig.~\ref{fig:back_door}. Mathematically, it computes the causal effect of the features $F$ to the predictions $Y$ at each stratum of the confounder $\mathcal{D}^p$, \ie,
\begin{align}
    p(Y|do(F))=\sum_d p(Y|F, \mathcal{D}^p=d)p(\mathcal{D}^p=d),
\end{align}
where the $do(\cdot)$ operation denotes %intervenes 
the estimation of the true causal effect via intervention, under the premise that the prediction of $p(Y|F)$ given each stratum $d$ of $\mathcal{D}^p$ is not biased. Unfortunately, this backdoor adjustment method is not applicable to our problem, as the pre-trained dataset $\mathcal{D}^p$ is oftentimes too huge to access and stratify.
}

\subsection{Front-door adjustment}
\ying{
Different from backdoor adjustment, front-door adjustment intervenes by introducing a mediator $Z$ in the forward path $F\rightarrow Y$, leading to the prediction
\begin{align}
    p(Y|F) = \sum_z p(z|F)p(Y|Z=z).
    \label{eqn:y_f}
\end{align}
In this case, deconfounding $\mathcal{D}^p$ requires both $F\rightarrow Z$ and $Z\rightarrow Y$ to be estimated with the true causal effect. 
First, $p(Z|do(F))=p(Z|F)$ since $Y$ works as a collider that blocks the information from $F$ to $Z$, \ie, $F\leftarrow \mathcal{D}^p\rightarrow Y\leftarrow Z$.
Second, similar to the stratification in back-door adjustment, the true causal effect $p(Y|do(Z=z))$ is obtained by computing at each stratum of $F$, \ie,
\begin{align}
    p(Y|do(Z=z)) = \sum_f p(Y|z,f)p(f).
    \label{eqn:y_doz}
\end{align}
Substituting Eqn.~(\ref{eqn:y_doz}) into Eqn.~(\ref{eqn:y_f}) gives the overall front-door adjustment as,
\begin{align}
    p(Y|do(F)) = \sum_z p(z|F) \sum_f p(Y|z,f)p(f).
\end{align}
}

\section{Pseudo-code}

We present the pseudo-codes of resolving rare features in Algorithm~\ref{alg:rare_features} and of resolving  spuriously correlated features in Algorithm~\ref{alg:spurious_features}.

\begin{algorithm}[h]
	\caption{Rare features.}
	\label{alg:rare_features}
	\begin{algorithmic}[1]
		\REQUIRE Pre-trained feature extractor $f$;  classification matrix $W$; queues $Q$; patch size $PS$; batch size $B$; iterations $T$; momentum m; the query parameters $\theta_q$ of $f$ and $W$.
		\STATE Initialize the momentum-updated model ($f_k$ and $W_k$) with parameters  $\theta_k$ by copying $\theta_q$.
		\STATE Randomly initialize the keys $Q_{y}$  for each category $y$.
		\FOR{iteration = $1$ to $T$} 
			\STATE Sample a batch of data $\left\{ \left( x_i, y_i 	\right)\right\}_{i=1}^{B}$;
			\STATE Generate image-level features $F_i = f(x_i)$ and predictions $W(F_i)$ for each image;
   			\STATE Get the confusing classes $y'_i$ and the ground true $y_i$;
   			\STATE Get 9 channels with $W_{y_i} - W_{y'_i}$ closest to zero and crop patches with a size
                     of $PS$ at the most attentive position of each selected channel;

			\STATE Obtain rare features $\left\{F_i^{r1}, F_i^{r2} ... F_i^{r9} \right\}$ from the cropped patches via  feature extractor $f$;
			\STATE Copy keys from $Q$ and detach.
			\FOR{$i$ = $1$ to $B$} 
				\STATE Obtain positive keys $F_{j}^{r}$ from $Q_{y_i}$;
				\STATE Obtain negative keys $F_{k}^{r}$ from $Q_{y \in y \ne y_i}$;
				\STATE Calculate losses for this sample: $\mathcal{L}_{r}^{i}$;
				\STATE Update queue: repeat $5-8$ using $f_k$ and $W_k$ to update rare features $F_{j}^{r}$ in $Q_{y_i}$;
			\ENDFOR
			\STATE  Average losses in the batch;
			\STATE  loss.backward();
   			\STATE  Update $\theta_q$ by gradients and update $\theta_k$ via $\theta_{k} \gets m\theta_{k} + (1-m)\theta_{q}$;
		\ENDFOR
	\end{algorithmic}
\end{algorithm}

\begin{algorithm}[h]
	\caption{Spuriously correlated features.}
	\label{alg:spurious_features}
	\begin{algorithmic}[1]
		\REQUIRE Pre-trained feature extractor $f$; patch/channel attention modules with aggregation operations  $\varphi_2$; classifier $\varphi_1$; batch size $B$; iterations $T$.
		\STATE Randomly initialize the $\varphi_1$ and $\varphi_2$.
		\FOR{iteration = $1$ to $T$} 
			\STATE Sample a batch of data $\left\{ \left( x_i, y_i 	\right)\right\}_{i=1}^{B}$;
			\STATE Generate mediator  $\hat{z}_i = \varphi_{2}(f(x_i))$ for each image;
			\STATE Obtain predictions  $\varphi_{1}(\hat{z}_i)$ for each image;
			\STATE Calculate losses for each sample: $\mathcal{L}_{s}^{i}$ ;
			\STATE  Average losses in the batch;
			\STATE  loss.backward();
		\ENDFOR
	\end{algorithmic}
\end{algorithm}

\section{Diagrams}
\subsection{Diagram of rare features}

\begin{figure*}[!h]
	\centering
	\includegraphics[width=0.7\linewidth]{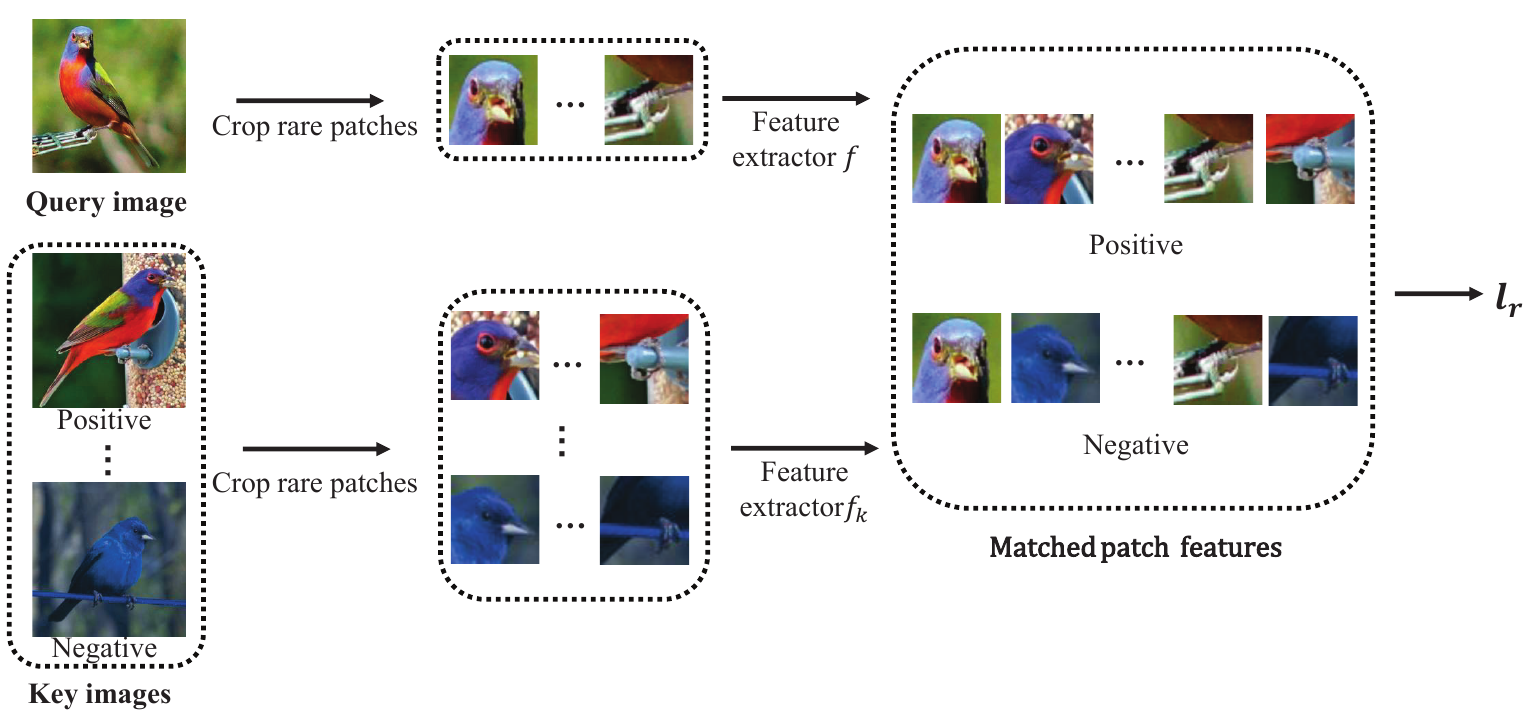}
        \caption{Illustration of resolving rare features. Given a query image and a set of key images, whose label is the same as the query image, are positive samples and otherwise are negative samples,  we first slice the images into rare patches based on the attentive regions of selected channels. Then the rare patch features are matched by EMD and used for calculating $\mathcal{L}_{r}$.}
	\label{fig:branch_2}
\end{figure*}

The detailed structure of resolving rare features is shown in Figure~\ref{fig:branch_2}.

\subsection{Diagram of spuriously correlated features} 
\begin{figure*}[!h]
	\centering
	\includegraphics[width=0.7\linewidth]{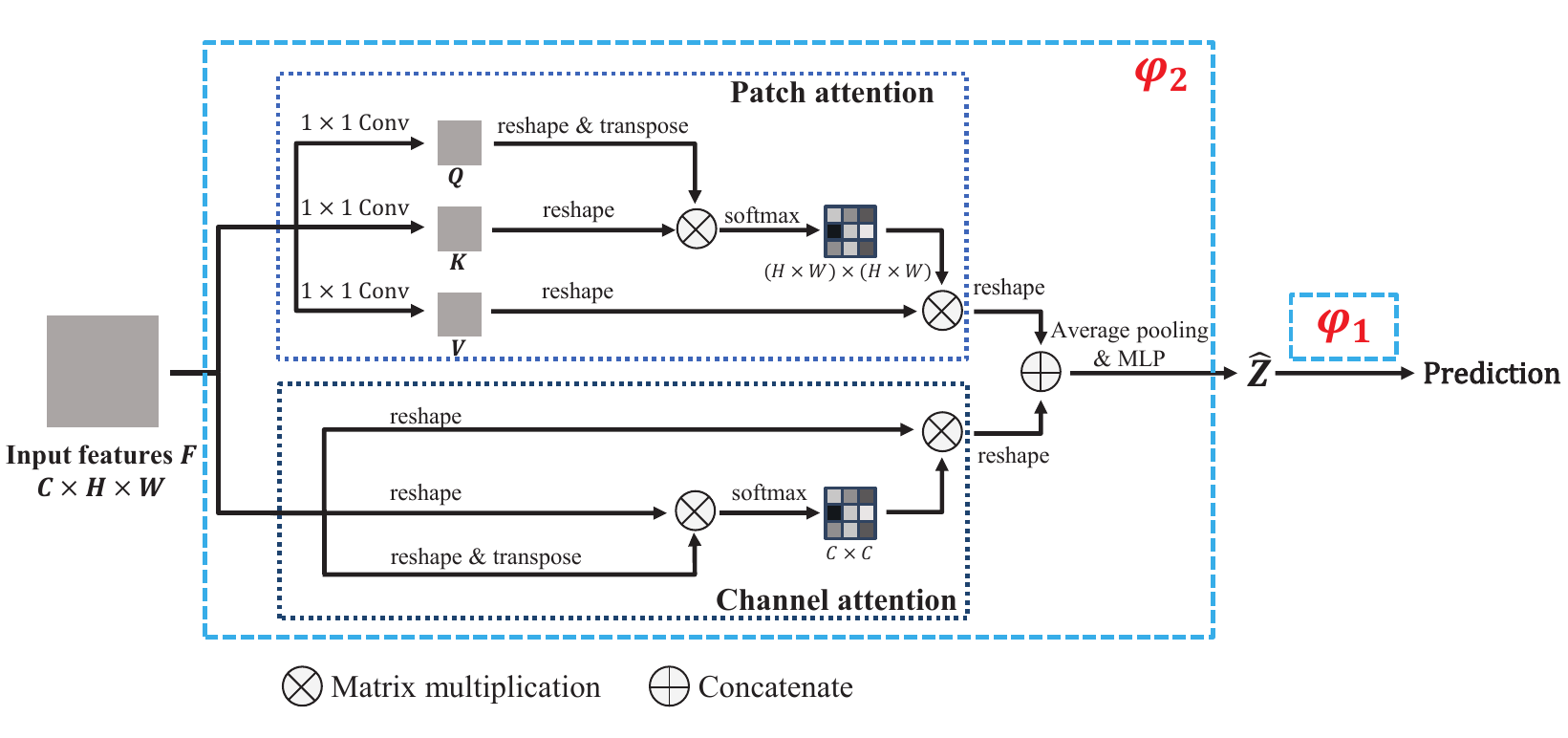}
	\caption{Illustration of resolving spuriously correlated features. Given an extracted feature map, we first use two attention modules to approximate the expectations and then aggregate them via a concatenation operation. Then, $\hat{Z}$ is obtained by an MLP, which is later used for predictions with $\varphi_{1}$.} 
	\label{fig:branch_3}
\end{figure*}

The detailed structure of resolving spuriously correlated features is shown in Figure~\ref{fig:branch_3}.

\section{Experimental Setup}
\subsection{Dataset description} 
Following~\cite{you2020co,zhong2020bi,zhang2021unleashing}, we evaluate our methods on eight datasets: \textbf{CUB-200-2011}~\cite{wah2011caltech}, \textbf{Stanford-Cars}~\cite{krause20133d}, \textbf{FGVC Aircraft}~\cite{maji2013fine}, \textbf{CIFAR10}, \textbf{CIFAR100}~\cite{krizhevsky2009learning}, \textbf{Vegetable}~\cite{Hou2017VegFru},  \textbf{ISIC}~\cite{codella2019skin} and \textbf{Caltech101}~\cite{fei2004learning}. In addition,  \textbf{ISIC} is a medical dataset with a highly imbalanced class distribution, which is closer to real-world applications. All datasets are obtained from the official websites agreeing with their licenses. The split of $100\%$ sampling rates either follows previous works~\cite{you2020co,zhong2020bi,zhang2021unleashing} except   {ISIC}. {ISIC} is a closed challenge dataset that does not provide the label of the official testing set. Thus we randomly split the official training set into our training set and testing set. Besides, the split in experiments of different data sizes (\ie, sampling rates $50\%$, $30\%$, and $15\%$)  is provided by previous works~\cite{you2020co,zhong2020bi}. We will also provide detailed image lists for the data split in our codes. Table~\ref{table:data_stastic} reports the statistics of all datasets.

\begin{table}[htbp]
	\caption{Statistics of datasets.}
	\label{table:data_stastic}
	\centering
		
	\begin{tabular}{lccc} 
		\toprule
		Datasets & Classes & Training & Testing \\
		\hline
		\textbf{CUB-200-2011} & 200 & 5994 & 5794 \\
		\textbf{Stanford-Cars} & 196 & 8144 & 8041 \\
		\textbf{FGVC Aircraft} & 100 & 6667 & 3333 \\
		\textbf{CIFAR10} & 10 & 50,000 & 10,000 \\
		\textbf{CIFAR100} & 100 & 50,000 & 10,000 \\
		\textbf{Vegetable} & 200 & 20,000 & 61,117 \\	
		\textbf{ISIC} & 7 & 5005 & 5010 \\	
		\textbf{Caltech101} & 102 & 3,060  & 6,084 \\
		\bottomrule
	\end{tabular}
%}
\end{table}

\begin{table*}[htbp]
	\caption{Statistics of hyper-parameters.}
	\label{table:hyper-parameters}
	\centering
	\resizebox{0.9\textwidth}{!}{
	\begin{tabular}{l|cccccccc} 
		\hline
		Hyper-parameters  & \textbf{CUB}      & \textbf{Cars}   & \textbf{Aircraft}   & \textbf{CIFAR10}  & \textbf{CIFAR100}   & \textbf{Vegetable}    & \textbf{ISIC}   & \textbf{Caltech101}    \\ 
		\hline
		Epochs & $50$ & \multicolumn{4}{|c|}{$20$} &  \multicolumn{3}{c}{$50$}\\
		\hline
		Iterations per epoch & \multicolumn{3}{c}{$500$} & \multicolumn{2}{|c|}{$200$} &  \multicolumn{3}{c}{$500$}  \\
		\hline
		Batch size & \multicolumn{3}{c}{$48$} & \multicolumn{2}{|c|}{$128$} &  \multicolumn{3}{c}{$48$}  \\
		\hline
		Learning rate (LR) & \multicolumn{8}{c}{$0.01$}  \\
		\hline
		LR schedule & MultiStep & \multicolumn{4}{|c|}{CosineAnnealing} &  \multicolumn{3}{c}{MultiStep}  \\
		\hline
		$K$ &\multicolumn{8}{c}{$40$}  \\
		\hline
		  Dimension of $\hat{Z}$ &\multicolumn{8}{c}{$512$}  \\
           \hline
		Patch size &\multicolumn{8}{c}{$64$}   \\
            \hline
  		Trade-off $\alpha$ &\multicolumn{8}{c}{$1.0$}   \\
            \hline
            Trade-off $\beta$  &\multicolumn{8}{c}{$5e-3$}   \\
		% \hline
  		\bottomrule
	\end{tabular}
	}
\end{table*}

\subsection{Environments and implement details} 
All methods are implemented in PyTorch~\cite{NEURIPS2019_9015} and on a computational platform with 8 Tesla V100 GPUs. Checkpoints of ResNet-50/ResNet-101 supervised pre-trained on ImageNet-1k are provided by PyTorch, and those of ResNet-50 self-supervised pre-trained on ImageNet-1k are obtained on their official implementation websites. Especially, SimCLR~\cite{chen2020simple} and BYOL~\cite{grill2020bootstrap} provide the checkpoints implemented in TensorFlow and we convert them into PyTorch using the codes approved by their authors.  For all datasets, we follow the data augmentations used in ~\cite{you2020co,zhong2020bi}: during training, images are randomly resized and cropped into $224 \times 224$ and then randomly horizon-flipped; during inference, images are resized to $256$ and then center-cropped to $224 \times 224$.  To be consistent with~\cite{zhong2020bi}, the batch size is set as $48$ for all datasets except {CIFAR10} and {CIFAR100}, and all methods are optimized by stochastic gradient descent with momentum $0.9$.  More statistics of hyper-parameters are reported in Table~\ref{table:hyper-parameters}.

\begin{table}[htbp]
	\caption{Percentages of testing images of CUB, on which the model trained from scratch makes correct predictions while fine-tuning methods misclassifies.}
	\label{table:mis_percents}
	\centering
        \resizebox{0.9\columnwidth}{!}{
	\begin{tabular}{lccc} 
		\toprule
		Dataset & Vanilla fine-tuning & Bi-tuning & Concept-tuning \\
		\hline
		CUB  & 3.08 & 2.74 &  \textbf{2.27} \\
		\bottomrule
	\end{tabular}
        }
	\vspace{-0.1in}
\end{table}

\section{Additional Experimental Results}			

\subsection{Compared to the model trained from scratch} 
In the appendix, we further report the comparison of the model trained from scratch and Concept-tuning. We choose the models on CUB using supervised pre-trained ResNet-50.  As shown in Table~\ref{table:mis_percents}, only $2.27\%$ images exist where the model trained from scratch makes correct predictions while Concept-tuning misclassifies, much better than Bi-tuning.

\subsection{Full table of results} 
\final{Table~\ref{table:pre-trained-vit} shows the experimental results using ViT/B-16 pre-trained by MAE and MoCo v3. Besides, in} the main paper we report the experimental results of five methods on three datasets using ResNet-50 pre-trained by different pre-training methods. We further report other methods in Table~\ref{tabel:trategies}. 

\begin{table}[!hbt] 
	\caption{{Results on fine-tuning ViT/B-16 by MAE and MoCo v3.}}
	\label{table:pre-trained-vit}
	\centering
	\resizebox{\columnwidth}{!}{
	\begin{tabular}{llccc}
		\toprule

  	  {Pre-training approach} &  Fine-tuning method    & CUB    & Car    & Aircraft     \\ 
  		\hline
		\multirow{4}{*}{\textbf{MAE}}    
		& Vanilla fine-tuning    & $77.80$ & $87.35$ & $86.98$  \\
		& Bi-tuning   			 & $79.29$ & $88.97$ & $87.91$ \\
		& Core-tuning            & $79.96$ & $89.54$ & $88.00$  \\
            &  \textbf{Ours  2}                & $\textbf{80.62}$ & $\textbf{90.05}$  & $\textbf{90.44}$  \\
		\hline
		\multirow{4}{*}{\textbf{MoCo v3}}    
		& Vanilla fine-tuning    & $80.34$ & $83.51$ & $85.60$  \\
		& Bi-tuning   			 & $83.76$ & $87.70$ & $88.45$ \\
		& Core-tuning            & $81.00$ & $ {89.52}$ & $89.31$  \\
            & \textbf{Ours 2}                & $\textbf{84.57}$ & $\textbf{90.91}$  & $\textbf{90.89}$  \\
		\bottomrule
		\end{tabular}
	}
\end{table}

\begin{table*}[hbtp] 
	\caption{Top-1 accuracy (\%) on three datasets using four different pre-trained ResNet-50. $*$ denotes that methods can not converge.}
	\label{tabel:trategies}
	\centering
	\resizebox{0.75\textwidth}{!}{
	\begin{tabular}{llccccc}
		\toprule
		\multirow{2}{*}{Dataset} & \multirow{2}{*}{Method} & \multicolumn{4}{c}{Pre-trained method}  \\
		\cmidrule{3-6}  
		&   & MoCo-V2      &  SimCLR    & SwAV   &  BYOL  &  Avg.    \\ 
		\hline
		\multirow{10}{*}{\textbf{CUB}}    
		& Vanilla fine-tuning    & $76.72 \pm 0.21$ & $76.51 \pm 0.28$ & $80.45 \pm 0.32$ & $81.29 \pm 0.29$ & $78.74$  \\
            & L2SP					 & $71.88 \pm 0.44$ & $64.55 \pm 0.53$ & $75.04 \pm 0.38$ & $76.72 \pm 0.25$ & $72.05$\\
            & DELTA					 & $72.87 \pm 0.38$ & $*$     & $*$     & $*$   &  $*$ \\
            & BSS					 & $76.92 \pm 0.29$ & $76.75 \pm 0.20$ & $80.89 \pm 0.35$ & $81.53 \pm 0.25$ & $79.02$\\
            & Co-tuning			     & $76.39 \pm 0.22$ & $76.35 \pm 0.17$ & $80.93 \pm 0.24$ & $81.72 \pm 0.31$ &   $78.85$\\
            & REGSL					 & $*$     &  $*$    & $77.70 \pm 0.29$ & $79.13 \pm 0.26$ &  $*$ \\
		& Bi-tuning   			 & $79.48 \pm 0.24$ & $75.73 \pm 0.25$ & $81.72 \pm 0.23$ & $82.02 \pm 0.29$  & $79.74$\\
		& Core-tuning            & $77.93 \pm 0.18$ & $77.55 \pm 0.15$ & $80.60 \pm 0.27$ & $78.46 \pm 0.18$  & $78.64$  \\
            &  Ours 1  & ${82.48} \pm 0.14$ & ${78.18}  \pm 0.20$ & ${83.47}  \pm 0.22$ & ${83.38}  \pm 0.18$  & $81.88$     \\
            &  Ours 2 & $\textbf{82.53} \pm \textbf{0.21}$ & $\textbf{79.81} \pm \textbf{0.23}$ & $\textbf{84.78} \pm \textbf{0.32}$ & $\textbf{84.45} \pm \textbf{0.29}$  & $\textbf{82.89}$   \\
		
		\hline
		\multirow{10}{*}{\textbf{Cars}}    
		& Vanilla fine-tuning     & $88.45 \pm 0.35$ & $84.53 \pm 0.12$ & $88.17 \pm 0.21$ & $88.99 \pm 0.39$  & $87.54$ \\  
            & L2SP					  & $81.58 \pm 0.28$ & $65.25 \pm 0.41$ & $76.51 \pm 0.27$ & $81.72 \pm 0.31$  &   $76.26$ \\
             & DELTA					  & $82.28 \pm 0.33$ & $*$     & $*$     & $*$      & $*$   \\
             & BSS					  & $88.07 \pm 0.27$ & $91.80 \pm 0.34$ & $88.07 \pm 0.31$ & $89.28 \pm 0.15$  & $89.31$   \\
             & Co-tuning			      & $88.35 \pm 0.16$ & $91.56 \pm 0.25$ & $88.53 \pm 0.17$ & $89.45 \pm 0.15$  & $89.47$ \\
             & REGSL					  & $90.96 \pm 0.19$ & $80.03 \pm 0.32$ & $78.60 \pm 0.24$ & $85.77 \pm 0.23$  & $83.84$ \\
		& Bi-tuning   			  & $90.05 \pm 0.15$ & $91.75 \pm 0.18$ & $90.49 \pm 0.27$ & $90.90 \pm 0.18$  & $90.80$\\	
		& Core-tuning             & $90.87 \pm 0.23$ & $91.78 \pm 0.26$ & $91.84 \pm 0.14$ & $91.95 \pm 0.18$  & $91.61$ \\
            &  Ours 1 & ${91.02} \pm 0.11$ & ${93.27} \pm 0.20$ & ${93.41} \pm 0.26$ & ${93.22} \pm 0.15$   & $92.73$ \\
            &  Ours 2 & $\textbf{91.75} \pm \textbf{0.18}$ & $\textbf{93.36} \pm \textbf{0.23}$ & $\textbf{93.79} \pm \textbf{0.32}$ & $\textbf{93.68} \pm \textbf{0.25}$  & $\textbf{93.15}$   \\

		\hline
		
		\multirow{10}{*}{\textbf{Aircraft}}   
		& Vanilla fine-tuning   & $88.60 \pm 0.18$ & $87.79 \pm 0.24$ & $83.26 \pm 0.17$ & $85.03 \pm 0.15$  & $86.17$ \\
            & L2SP					 & $86.17 \pm 0.25$ & $65.25 \pm 0.52$ & $76.27 \pm 0.42$ & $80.32 \pm 0.23$ & $77.00$ \\
            & DELTA					  & $82.28 \pm 0.33$ & $*$     & $*$     & $*$      &   $*$   \\
            & BSS					 & $88.63 \pm 0.17$ & $88.30 \pm 0.30$ & $83.59 \pm 0.15$ & $84.64 \pm 0.20$ & $86.29$\\
            & Co-tuning			     & $88.63 \pm 0.23$ & $87.64 \pm 0.35$ & $83.59 \pm 0.24$ & $84.52 \pm 0.16$ &$86.10$ \\
            & REGSL					 & $81.16 \pm 0.21$ & $*$     & $*$     & $73.90 \pm 0.37$ & $*$   \\
		& Bi-Tuning         	& $89.05 \pm 0.16$ & $88.69 \pm 0.17$ & $85.69 \pm 0.13$ & $87.16 \pm 0.11$  & $87.65$  \\
		& Core-Tuning   	 & $89.02 \pm 0.19$ & $89.47 \pm 0.21$ & $88.66 \pm 0.34$ & ${89.74} \pm {0.20}$  & $89.22$ \\
            & Ours 1  & $\textbf{89.65} \pm \textbf{0.18}$ & ${90.13} \pm 0.11$ & ${91.42} \pm 0.36$ & ${90.82} \pm 0.22$  & $90.50$ \\
            &  Ours 2  & ${89.32} \pm 0.21$ & $\textbf{90.85} \pm \textbf{0.17}$ & $\textbf{91.75} \pm \textbf{0.14}$ & $\textbf{91.21} \pm \textbf{0.13}$  & $\textbf{90.76} $   \\

		\bottomrule
		\end{tabular}
	}
\end{table*}

\subsection{More ablation studies}
\begin{figure}[hbt]
  \centering
    \begin{subfigure}[t]{0.49\columnwidth}
  \centering
  	\includegraphics[scale=0.15]{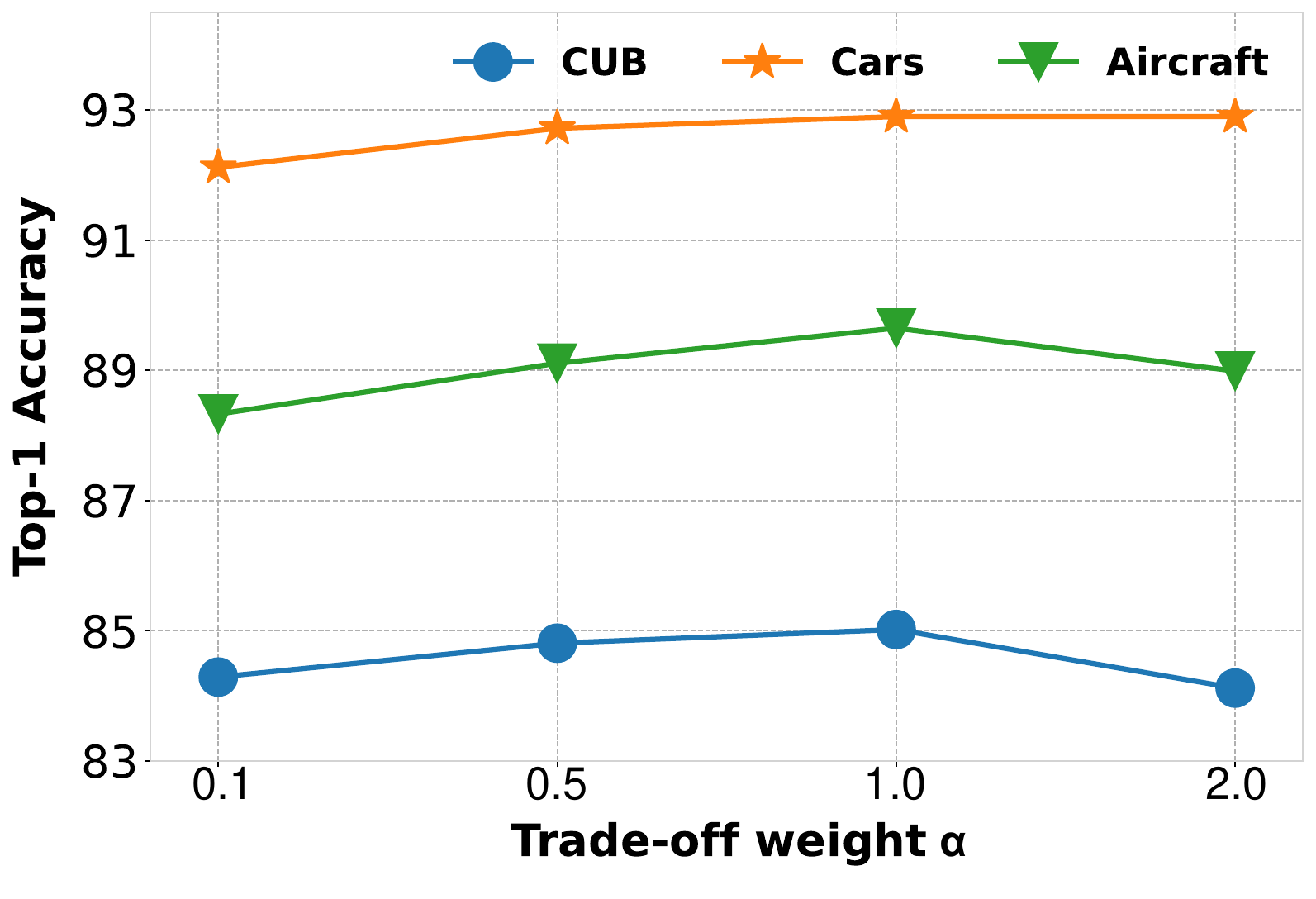}%}
  	\caption{}
    \label{fig:sen_lr_weights}
  \end{subfigure}
    \hfill
      \begin{subfigure}[t]{0.49\columnwidth}
  \centering
  	\includegraphics[scale=0.15]{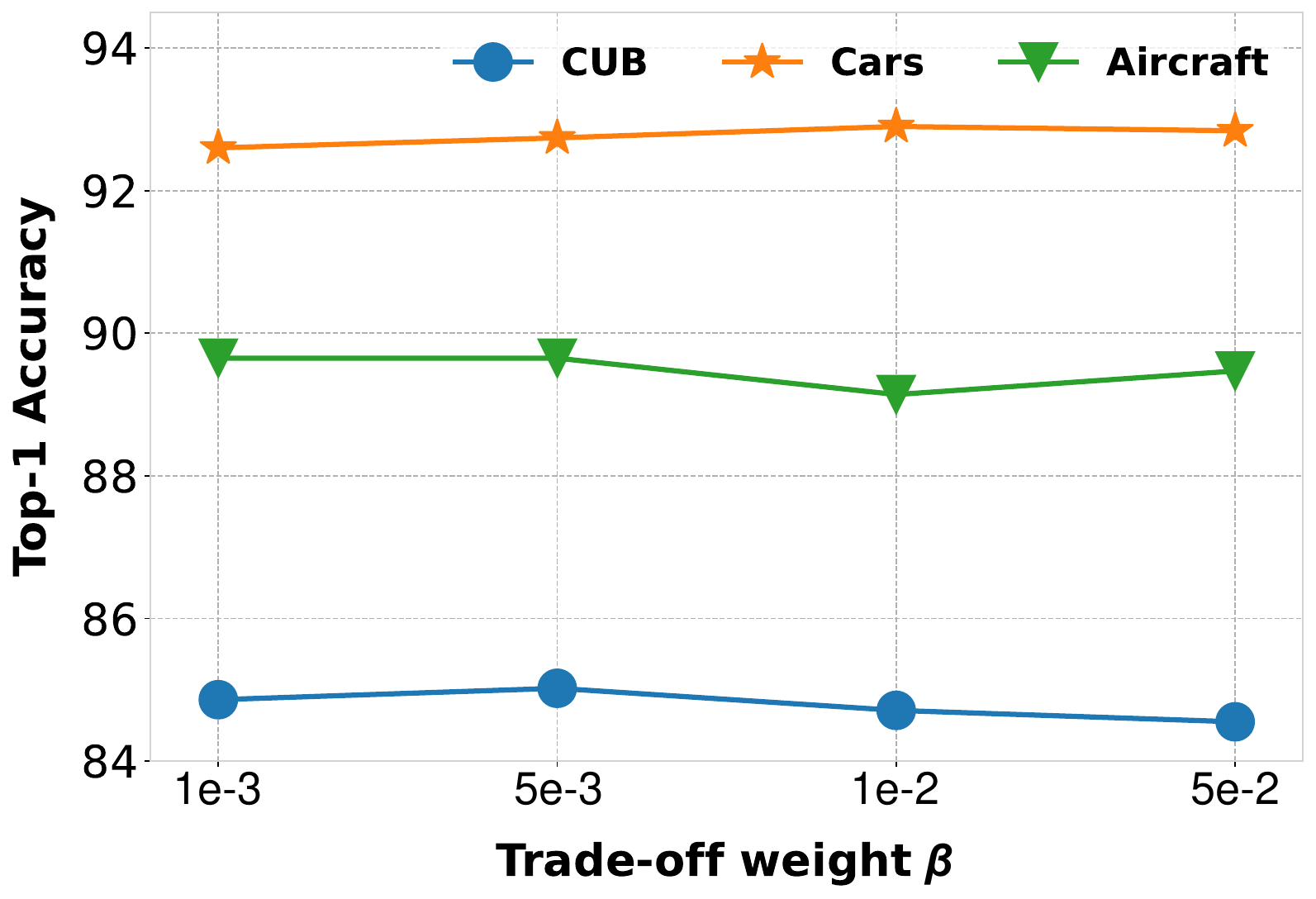}%}
  	\caption{}
    \label{fig:sen_ls_weights}
  \end{subfigure}
  \caption{Analysis of $\alpha$ and $\beta$ in our methods  using supervised pre-trained ResNet-50. }
  \vspace{-0.1in}
\end{figure}

\textbf{Influences of the trade-off weight $\alpha$}.
In previous experiments, we set the trade-off weights $\alpha$ as $1.0$ for all datasets. In this section, we further analyze how this term influences performance. Larger $\alpha$ will more strongly pull rare features and, in the meantime, will increase the risk of overfitting. The results on three datasets in Fig.~\ref{fig:sen_lr_weights} show clear peaks, which supports the existence of a trade-off.

\textbf{Influences of the trade-off weight} $\beta$. In previous experiments, we defaulted trade-off weights $\beta$ as $5e-3$ for all datasets. Here, we further analyze how this term influences our methods. We try different values on three datasets, and the results are shown in Fig.~\ref{fig:sen_ls_weights}.  Even though the best $\beta$ may vary on different datasets, the default value $5e-3$ is enough to obtain good performances.

 \begin{figure*}[!hbt]
	\begin{subfigure}[]{0.33\linewidth}
		\centering
		{\includegraphics[width = 0.9\linewidth]{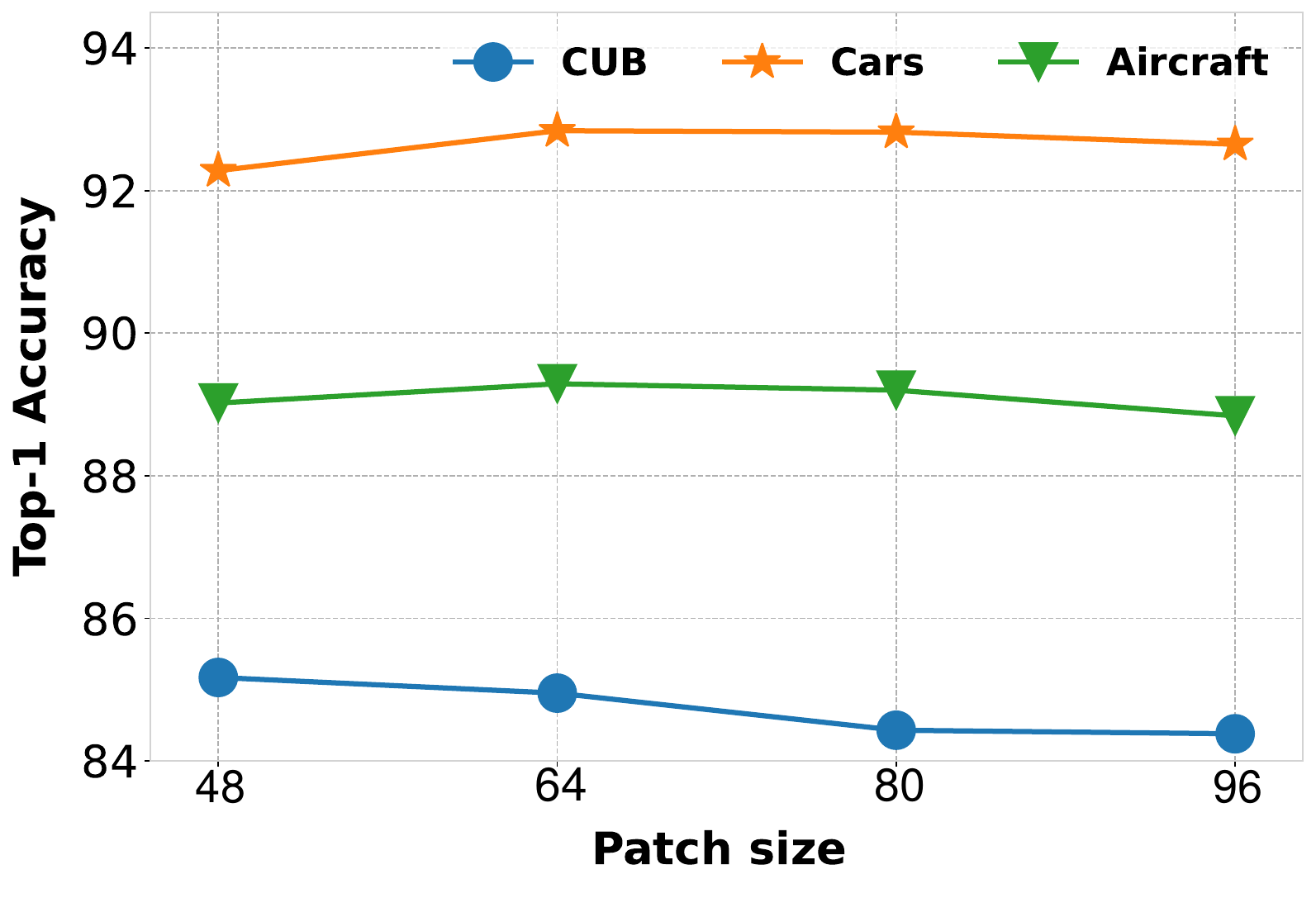}}
            % \vspace{-0.1in}
      	\caption{}
    	\label{fig:sen_patch_size}
	\end{subfigure}
	\begin{subfigure}[]{0.33\linewidth}
		\centering
		{\includegraphics[width =  0.9\linewidth]{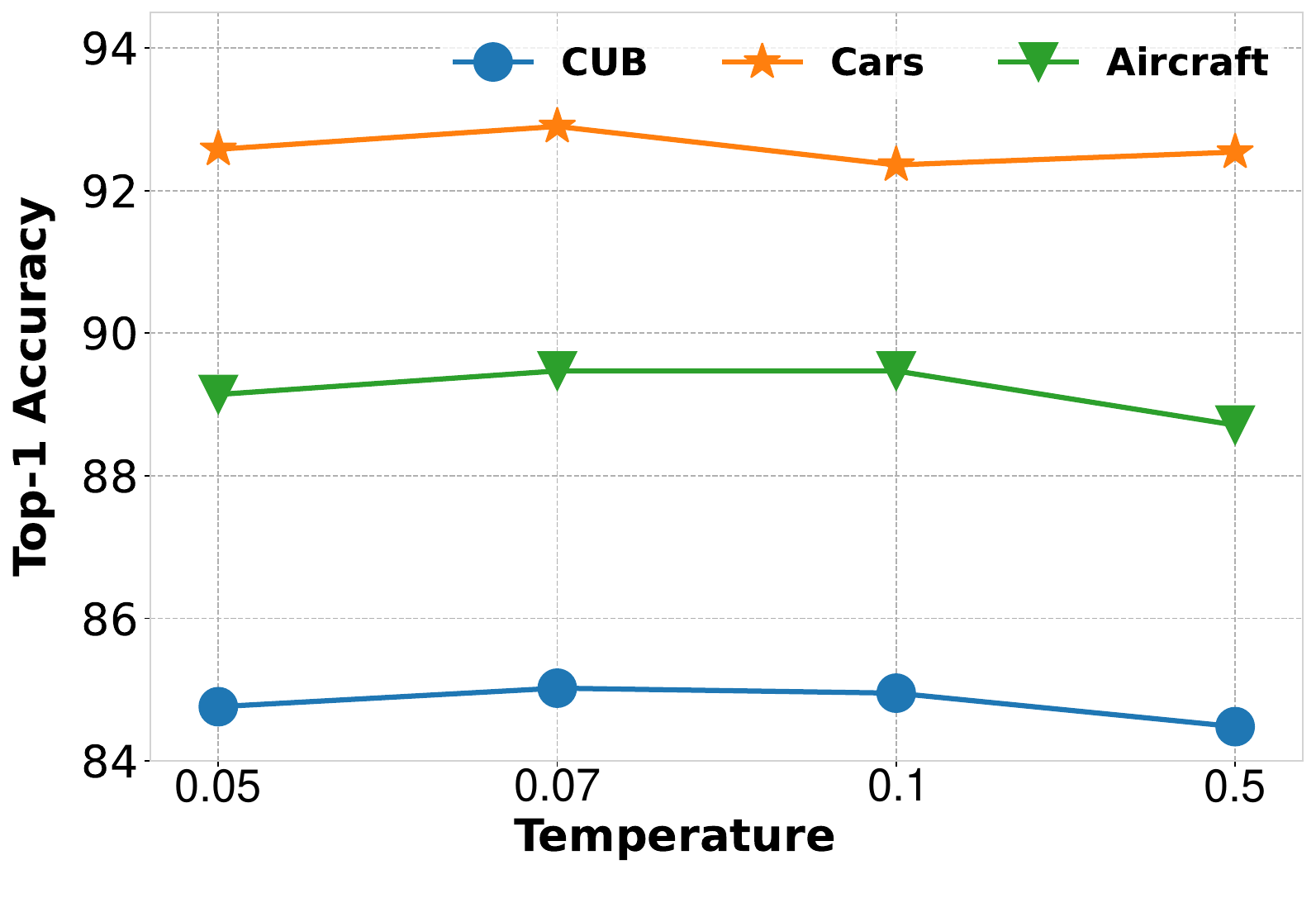}}
            % \vspace{-0.1in}
      	\caption{}
	       \label{fig:sen_t}
	\end{subfigure} 
	\begin{subfigure}[]{0.33\linewidth}
		\centering
    	{\includegraphics[width =  0.9\linewidth]{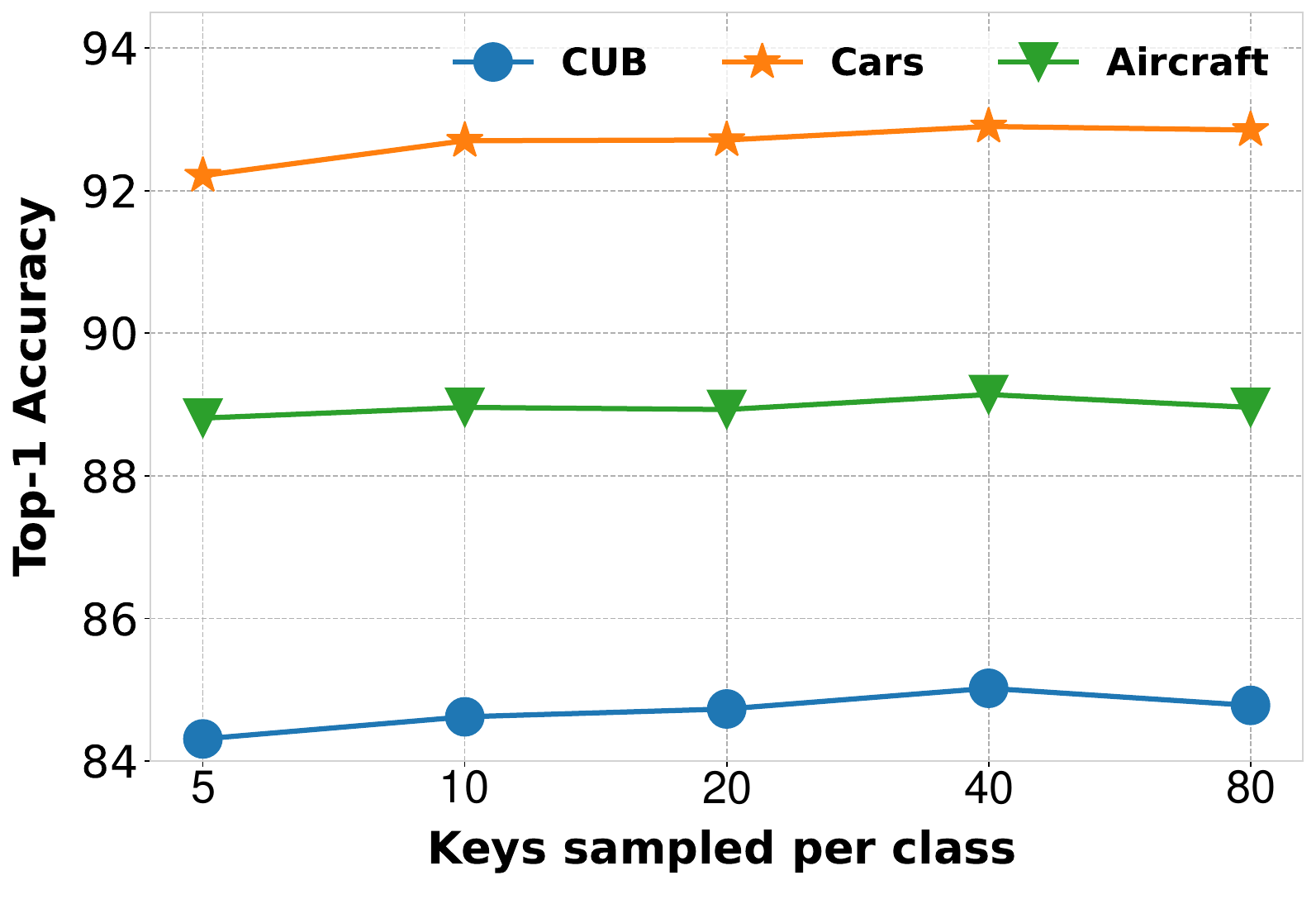}}
            % \vspace{-0.1in}
      	\caption{}
    	\label{fig:sen_keys}
	\end{subfigure}
 % \vspace{-0.1}
\caption{Analysis of the patch size, temperature, and keys in our methods on three datasets using supervised pre-trained  ResNet-50.}
\vspace{-0.2in}
\end{figure*}

\textbf{Influences of the patch size.}
\ying{Selection of an appropriate} patch size is also important in our methods and the best patch size varies \ying{from dataset to dataset. While} features in larger patch sizes contain more information and are more discriminative for contrastive learning,  \ying{a sufficiently large patch that occupies most regions of an object and likely fuses both rare and non-rare features decreases} the effectiveness of resolving rare features. For example, as shown in Fig.~\ref{fig:sen_patch_size}, the best patch size for {CUB}  is $48$ while the best one for {Cars} and {Aircraft} is $64$ because the objects in CUB occupy relatively smaller regions in the images. Furthermore, when the patch size increases to $96$, the performances will decrease a lot (\eg, performance drops from $85.17\%$ to $84.38\%$ on CUB).

\textbf{Influences of the temperature $\tau$.}
In previous experiments, we follow the implementation of supervised contrastive learning~\cite{khosla2020supervised} to set the temperature $\tau$ as $0.07$. In this section, we conduct experiments to explore the influences of $\tau$. As shown in Fig.~\ref{fig:sen_t}, $\tau$ as $0.07$ obtains better performances. The potential reason is that smaller $\tau$ tends to punish hard samples more~\cite {zhang2022dual} for generating more universal representations while reducing the tolerances of hard samples. 

\textbf{Influences of the number of keys.}
This section discusses the influences of the number of keys used in $\mathcal{L}_{r}$. In contrastive learning, a larger queue is beneficial~\cite{chen2020simple}. However, a large number of keys in supervised contrastive learning necessarily sacrifice the stochasticity of sampling from the queue~\cite{zhong2020bi}, especially under limited training data (\eg, up to $30$ training samples per class on CUB), unfavorably easing contrastive learning and leading to saturation. Our experimental results in Fig.~\ref{fig:sen_keys} present apparent saturation, which supports the unnecessary storage of too many keys. Note that the best number of keys varies in different datasets, and our default value of $40$ also performs well.

{\section{Additional Visualization Results}}
\subsection{More visualization of attentive regions}
\begin{figure}[!h]
  \centering
  \includegraphics[width=0.8\linewidth]{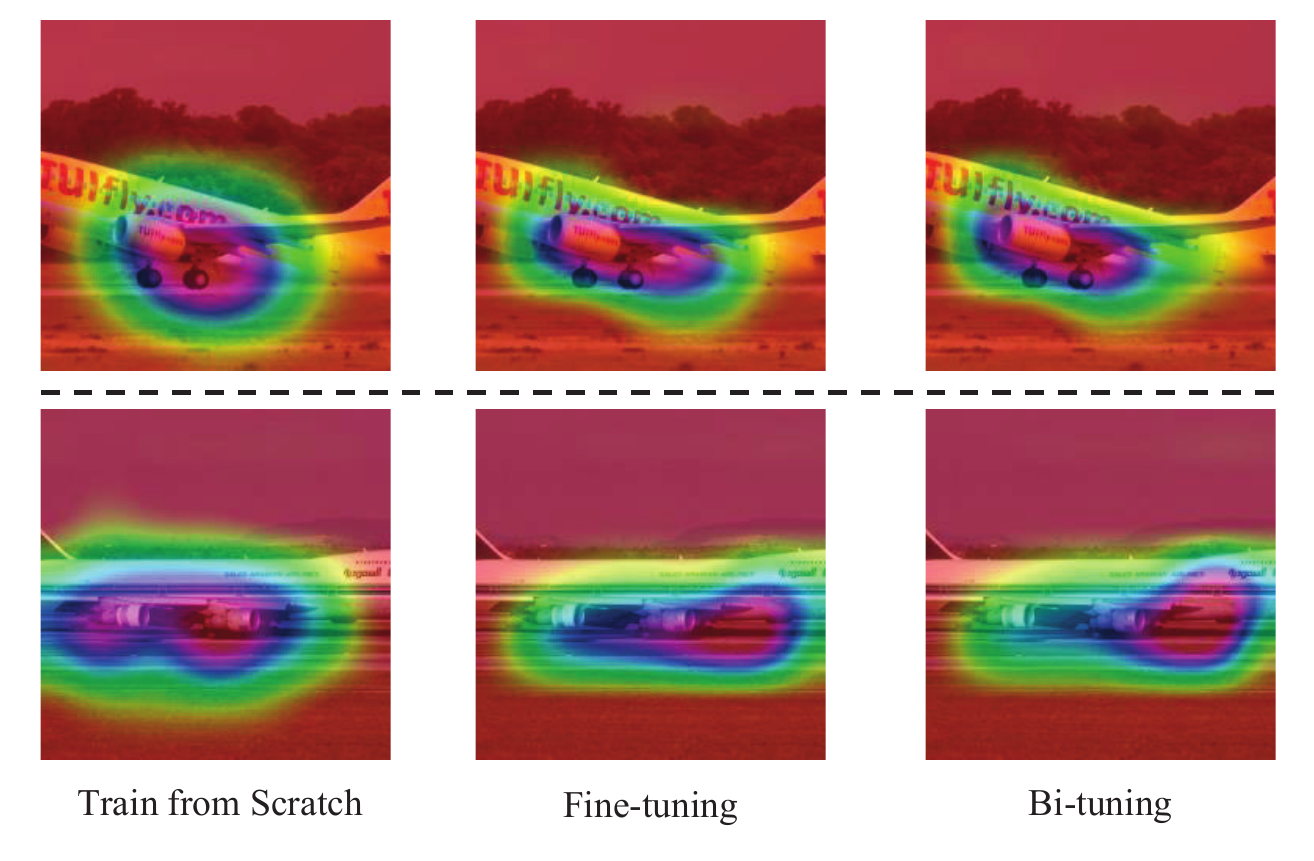}
  \caption{Exemplar attentive regions of the model trained \ying{ (a) from scratch, by (b) fine-tuning and (c) bi-tuning, where only the first column predicts correctly.}} %, respectively.}
  \label{fig:Aircraft_cam}
  \end{figure}

In the section, We further show the regions attended by the three models on \textbf{Aircraft} in Figure~\ref{fig:Aircraft_cam}. Consistent with the difference in predictions, the two fine-tuning methods rely more on front power-plant features.

\begin{figure}[!h]
  \centering
  \includegraphics[width=0.9\linewidth]{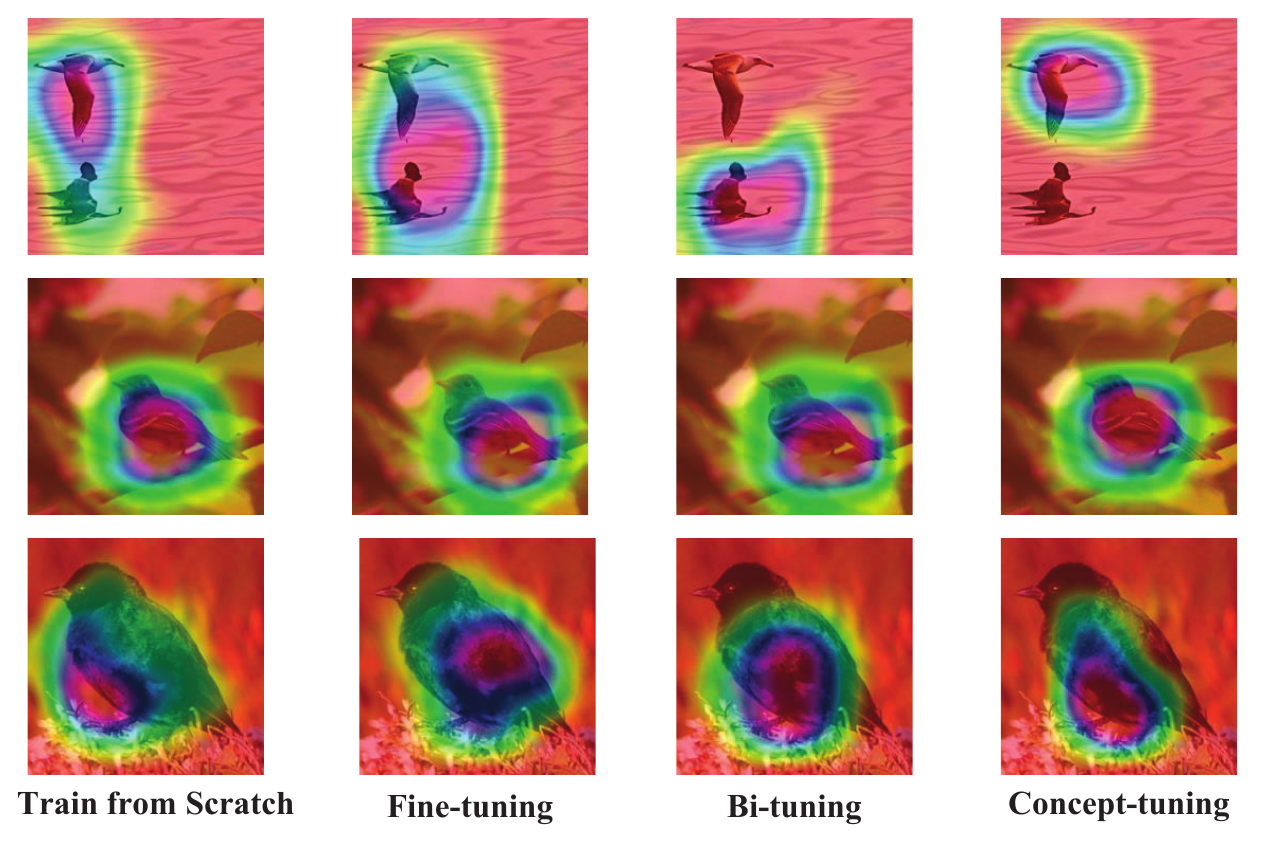}
  \caption{CAM visualization of four methods.} 
  \label{fig:ours_cam_supp}
  \end{figure}

\subsection{More visualization of Concept-tuning}
To better understand the effectiveness of our methods, we provide several examples of CAMs in different methods, as shown in Fig.~\ref{fig:ours_cam_supp}. Influenced by the pre-trained model, fine-tuning and Bi-tuning will be attracted by the pre-training features and make wrong predictions, while Concept-Tuning could resolve the negative effects and predict correctly. For example, fine-tuning and Bi-tuning neglect the chest features as shown in the second row of Fig.~\ref{fig:ours_cam_supp}, while Concept-tuning attends regions closer to the model trained from scratch.

\end{document}